\documentclass{article}

\usepackage{microtype}
\usepackage{graphicx}
\usepackage{subfigure}
\usepackage{booktabs} 

\usepackage{xcolor}
\usepackage{bm}
\usepackage{multirow}

\newcommand{\R}{\mathbb{R}}
\newcommand{\diff}{\mathop{}\!\mathrm{d}}
\newcommand{\expp}{\mathrm{e}}
\newcommand{\cond}{{\;|\;}}

\usepackage{xcolor, soul}
\sethlcolor{pink}

\newcommand{\parsection}[1]{\textbf{#1 }}

\makeatletter
\def\changegreek{\@for\next:={%
  alpha,beta,gamma,delta,epsilon,zeta,eta,theta,kappa,lambda,mu,nu,xi,pi,rho,sigma,%
  tau,upsilon,phi,chi,psi,omega,varepsilon,vartheta,varpi,varrho,varsigma,varphi}%
  \do{\expandafter\let\csname\next\expandafter\endcsname\csname\next up\endcsname}}
\def\changegreekbf{\@for\next:={%
  alpha,beta,gamma,delta,epsilon,zeta,eta,theta,kappa,lambda,mu,nu,xi,pi,rho,sigma,%
  tau,upsilon,phi,chi,psi,omega,varepsilon,vartheta,varpi,varrho,varsigma,varphi}%
  \do{\expandafter\def\csname\next\expandafter\endcsname\expandafter{%
    \expandafter\bm\expandafter{\csname\next up\endcsname}}}}
\makeatother

\usepackage{hyperref}

\usepackage[accepted]{icml2023}

\usepackage{amsmath}
\usepackage{amssymb}
\usepackage{mathtools}
\usepackage{amsthm}

\usepackage[capitalize,noabbrev]{cleveref}

\theoremstyle{plain}
\newtheorem{theorem}{Theorem}[section]
\newtheorem{proposition}[theorem]{Proposition}

\theoremstyle{definition}

\theoremstyle{remark}

\usepackage[textsize=tiny]{todonotes}

\icmltitlerunning{Image Restoration with Mean-Reverting Stochastic Differential Equations}

\begin{document}

\twocolumn[

\icmltitle{Image Restoration with Mean-Reverting Stochastic Differential Equations}



\icmlsetsymbol{equal}{*}

\begin{icmlauthorlist}
\icmlauthor{Ziwei Luo}{uu}
\icmlauthor{Fredrik K. Gustafsson}{uu}
\icmlauthor{Zheng Zhao}{uu}
\icmlauthor{Jens Sj{\"o}lund}{uu}
\icmlauthor{Thomas B. Sch{\"o}n}{uu}
\end{icmlauthorlist}

\icmlaffiliation{uu}{Department of Information Technology, Uppsala University, Sweden}

\icmlcorrespondingauthor{Ziwei Luo}{ziwei.luo@it.uu.se}

\icmlkeywords{Image Restoration, Stochastic Differential Equations, Diffusion model}

\vskip 0.3in
]



\printAffiliationsAndNotice{}  

\begin{abstract}
    This paper presents a stochastic differential equation (SDE) approach for general-purpose image restoration. The key construction consists in a mean-reverting SDE that transforms a high-quality image into a degraded counterpart as a mean state with fixed Gaussian noise. Then, by simulating the corresponding reverse-time SDE, we are able to restore the origin of the low-quality image without relying on any task-specific prior knowledge. Crucially, the proposed mean-reverting SDE has a closed-form solution, allowing us to compute the ground truth time-dependent score and learn it with a neural network. Moreover, we propose a maximum likelihood objective to learn an optimal reverse trajectory that stabilizes the training and improves the restoration results. The experiments show that our proposed method achieves highly competitive performance in quantitative comparisons on image deraining, deblurring, and denoising, setting a new state-of-the-art on two deraining datasets. Finally, the general applicability of our approach is further demonstrated via qualitative results on image super-resolution, inpainting, and dehazing. Code is available at \url{https://github.com/Algolzw/image-restoration-sde}.

\end{abstract}

\section{Introduction}
\label{section:introduction}

Diffusion models have shown impressive performance in various image generation tasks, based on modeling a diffusion process and then learning its reverse~\citep{sohl2015deep, ho2020denoising, song2019generative, song2020improved, song2021denoising, song2021maximum, song2021score, rombach2022high,rissanen2022generative}. Among the commonly used formulations \citep{yang2022diffusion}, we adopt that of using the diffusion models defined via stochastic differential equations~\citep[SDEs,][]{song2021maximum,song2021score}. This entails gradually diffusing images towards a pure noise distribution using an SDE, and then generating samples by learning and simulating the corresponding reverse-time SDE \citep{anderson1982reverse}. The essence is training a neural network to estimate the score function of the noisy data distributions \citep{song2019generative}.
\looseness=-1

Image restoration is the general task of restoring a high-quality image from a degraded low-quality version. Common specific examples include image deraining \citep{li2019single, ren2019progressive}, deblurring \citep{nah2017deep, zhang2020deblurring}, denoising \citep{zhang2017beyond, zhang2018ffdnet}, and super-resolution \citep{dong2015image, lugmayr2020srflow,luo2022deep}, just to mention a few.
Image restoration has a rich history \citep{hunt1973application, andrews1974digital, sezan1990survey, banham1997digital} and remains an active topic within computer vision where learning-based approaches have a prominent role \citep{zhang2017image, zhang2017learning, wang2022uformer, xiao2022stochastic}. 

\begin{figure*}[t]
    \begin{center}
    \centerline{\includegraphics[width=1.\textwidth]{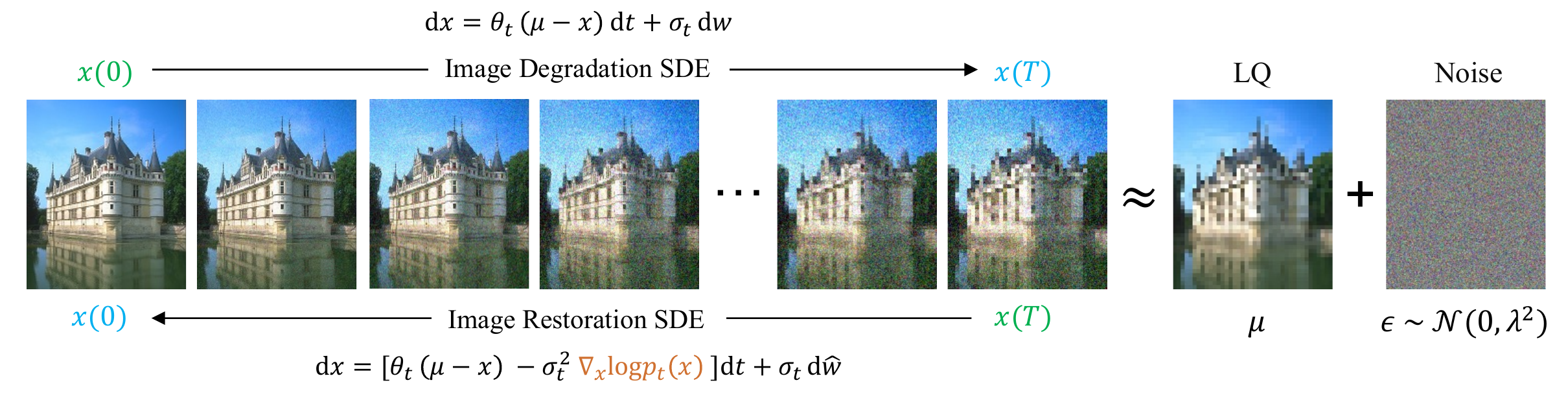}}\vspace{-2.0mm}
    \caption{An overview of our proposed construction, where a mean-reverting SDE (\ref{equ:ou}) is used for image restoration. The SDE models the degradation process from a high-quality image $x(0)$ to its low-quality counterpart $\mu$, by diffusing $x(0)$ towards a noisy version $\mu + \epsilon$ of the low-quality image. By simulating the corresponding reverse-time SDE, high-quality images can then be restored.}
    \label{fig:overview}
    \end{center}
    \vskip -0.2in
\end{figure*}

Diffusion models have recently been applied to different image restoration tasks. \citet{saharia2022image, saharia2022palette} train diffusion models which are conditioned on the low-quality images, while \citet{lugmayr2022repaint} utilize a pretrained unconditional model together with a modified generative process. Others explicitly treat image restoration as an inverse problem, assuming that the degradation and its parameters are known at test-time \citep{kawar2021snips, chung2022diffusion, kawar2022denoising}. These methods all employ the standard forward process, which diffuses images to pure noise. The reverse (generative) processes are thus initialized with sampled noise of high variance, which can result in poor restoration of the ground truth high-quality image. A number of experiments have shown that diffusion models can produce better perceptual scores, but often perform unsatisfactory in terms of some pixel/structure based distortion criteria~\citep{saharia2022image, li2022srdiff, kawar2021snips}.
\looseness=-1

To address this issue, we propose to solve the image restoration problem using a \emph{mean-reverting} SDE. As illustrated in Figure~\ref{fig:overview}, this adapts the forward process such that it models the image degradation itself, from a high-quality image to its low-quality counterpart. By simulating the corresponding reverse-time SDE, high-quality images can be restored. Importantly, no task-specific prior knowledge is required to model the image degradation at test time, just a set of image pairs for training. Our main contributions are as follows:
\begin{itemize}
    \item We propose a general-purpose image restoration approach using a mean-reverting SDE that directly models the image degradation process. Our formulation has a closed-form solution that enables us to compute the ground truth time-dependent score function and train a neural network to estimate it.

    \item We propose a simple alternative loss function for training the neural network, based on maximizing the likelihood of the reverse-time trajectory. The loss is demonstrated to stabilize training and consistently improve the image restoration performance compared to the common score matching objective.

    \item We demonstrate the general applicability of our proposed approach by applying it to six diverse image restoration tasks: image deraining, deblurring, denoising, super-resolution, inpainting and dehazing.

    \item Our approach achieves highly competitive restoration performance in quantitative comparisons on image deraining, deblurring and denoising, setting a new state-of-the-art on two deraining datasets.
\end{itemize}

\section{Background}
\label{section:background}

In this section, we briefly review the key concepts underlying SDE-based diffusion models and show the process of generating samples with reverse-time SDEs.
Let $p_0$ denote the initial distribution that represents the data, and $t \in [0, T]$ denote the continuous time variable. We consider a diffusion process $\{{x}(t)\}_{t=0}^T$ defined by an SDE of the form,
\begin{equation}
	\diff {x} = f({x}, t) \diff t + g(t)\diff w, \quad {x}(0) \sim p_0({x}), 
	\label{equ:sde}
\end{equation}
where $f$ and $g$ are the drift and dispersion functions, respectively, $w$ is a standard Wiener process, and ${x}(0) \in \R^{d}$ is an initial condition. Typically, the terminal state ${x}(T)$ follows a Gaussian distribution with fixed mean and variance. The general idea is to design such an SDE that gradually transforms the data distribution into fixed Gaussian noise~\citep{song2021score,lu2022dpm,de2022riemannian}.

We can then reverse the process to sample data from noise by simulating the SDE backward in time~\citep{song2021score}. \citet{anderson1982reverse} shows that a reverse-time representation of the SDE~\eqref{equ:sde} is given by
\begin{equation}
    \begin{split}
        \diff {x} &= \Bigl[ f({x}, t) - g(t)^2\, \nabla_{{x}} \log p_t({x}) \Bigr] \diff t + g(t) \diff \hat{w},
    \end{split}
    \label{equ:reverse-sde}
\end{equation}
where ${x}(T) \sim p_T({x})$. Here, $\hat{w}$ is a reverse-time Wiener process and $p_t({x})$ stands for the marginal probability density function of ${x}(t)$ at time~$t$. The score function $\nabla_{x} \log p_t({x})$ is in general intractable and thus SDE-based diffusion models approximate it by training a time-dependent neural network $s_\theta({x}, t)$ under a so-called score matching objective~\citep{hyvarinen2005estimation,song2021score}.

\section{Method}
\label{section:method}

The key idea of our proposed image restoration approach is to combine a mean-reverting SDE with a maximum likelihood objective for neural network training. We thus refer to it as an \emph{Image Restoration Stochastic Differential Equation} (IR-SDE). We begin by describing the forward and reverse processes of the mean-reverting SDE, and adapt previously described, score-based, training methods to estimate this SDE. Then, we describe and contrast this with our proposed loss function based on a maximum likelihood objective.

\subsection{Forward SDE for Image Degradation}

We construct a special case of the SDE~\eqref{equ:sde} whose score function is analytically tractable, as follows:
\begin{equation}
	\diff {x} = \theta_t \, (\mu - {x}) \diff t + \sigma_t \diff w,
	\label{equ:ou}
\end{equation}
where $\mu$ is the state mean, and $\theta_t, \sigma_t$ are time-dependent positive parameters that characterize the speed of the mean-reversion and the stochastic volatility, respectively. There is a lot of freedom when it comes to choosing~$\theta_t$ and~$\sigma_t$ and, as we will see in Section~\ref{section:discussion:theta}, the choice can have a significant impact on the resulting restoration performance. 

In general, $\mu$ and the starting state ${x}(0)$ can be set to any pair of different images. The forward SDE (\ref{equ:ou}) then transfers one image to the other as a kind of noisy interpolation. To carry out image degradation, we let ${x}(0)$ and $\mu$ be the ground truth high-quality (HQ) image and its degraded low-quality (LQ) counterpart, respectively (see Figure~\ref{fig:overview}). It is worth noting that while $\mu$ thus depends on ${x}(0)$ (as they are paired HQ-LQ images of the same object or scene), ${x}(0)$ is independent of the Brownian motion and the SDE is therefore still valid in the It\^{o} sense.

For our SDE \eqref{equ:ou} to have a closed-form solution, we set $\sigma_t^2 \, / \, \theta_t = 2 \, \lambda^2$, where $\lambda^2$ is the stationary variance. With this, we have the following:

\begin{proposition}
    Suppose that the SDE coefficients in \eqref{equ:ou} satisfy $\sigma_t^2 \, / \, \theta_t = 2 \, \lambda^2$ for all times $t$. Then, given any starting state ${x}(s)$ at time $s < t$, the solution to the SDE is
    \begin{equation}
        {x}(t) = \mu + \bigl({x}(s) - \mu \bigr) \, \expp^{-\bar{\theta}_{s:t}} + \int^t_s \sigma_z \, \expp^{-\bar{\theta}_{z:t}} \diff w(z),
    \end{equation}
    where $\bar{\theta}_{s:t} \coloneqq \int^t_s \theta_z \diff z$ is known and the transition kernel $p({x}(t) \cond {x}(s)) = \mathcal{N}\bigl({x}(t) \cond m_{s:t}({x}(s)), v_{s:t}\bigr)$ is a Gaussian with mean $m_{s:t}$ and variance $v_{s:t}$ given by:
    \begin{equation}
        \begin{split}
        m_{s:t}(x(s)) &\coloneqq \mu + ({x}(s) - \mu) \, \expp^{-\bar{\theta}_{s:t}}, \\
        v_{s:t} &\coloneqq \int^t_s \sigma_z^2 \, \expp^{-2\bar{\theta}_{z:t}} \diff z \\
        & \ = \lambda^2 \, \Bigl(1 - \expp^{-2 \, \bar{\theta}_{s:t}}\Bigr).
        \end{split}
    \end{equation}
\label{prop:forward_sde_solution}
\end{proposition}

The proof is provided in Appendix~\ref{prf:SDE_solution}. To simplify the notation when the starting state is ${x}(0)$, we substitute $\bar{\theta}_{0:t}, m_{0:t}, v_{0:t}$ with $\bar{\theta}_{t}, m_{t}, v_{t}$, respectively. 
Then we have the distribution of ${x}(t)$ at any time $t$ conditioned on the initial state, given by
\begin{equation}
    \begin{split}
    p_t({x}) & \ = \mathcal{N}\bigl({x}(t) \cond m_{t}({x}), v_{t}\bigr), \\
    m_{t}({x}) &\coloneqq \mu + ({x}(0) - \mu) \, \expp^{-\bar{\theta}_{t}}, \\
    v_{t} & \coloneqq \lambda^2 \, \Bigl(1 - \expp^{-2 \, \bar{\theta}_{t}}\Bigr).
    \end{split}
    \label{eq:sde_gauss_mean_var}
\end{equation}
Note that as $t \to \infty$, the mean $m_{t}$ converges to the low-quality image $\mu$ and the variance $v_{t}$ converges to the stationary variance $\lambda^2$ (hence the qualifier ``mean-reverting''). In other words, the forward SDE~\eqref{equ:ou} diffuses the high-quality image into a low-quality image with fixed Gaussian noise.
\looseness=-1

\subsection{Reverse-Time SDE for Image Restoration}

To recover the high-quality image from the terminal state ${x}(T)$, we reverse the SDE~\eqref{equ:ou} according to \eqref{equ:reverse-sde} to derive an image restoration SDE (IR-SDE), given by
\begin{equation}
    \diff {x} = \big[ \theta_t \, (\mu - {x}) - \sigma_t^2 \, \nabla_{{x}} \log p_t({x}) \big] \diff t + \sigma_t \diff \hat{w}.
    \label{eq:reverse-irsde}
\end{equation}
At test time, the only unknown part is the score $\nabla_{{x}} \log p_t({x})$ of the marginal distribution at time $t$. But during training, the ground truth, high-quality image ${x}(0)$ is available and thus we can train a neural network to estimate the conditional score $\nabla_{{x}} \log p_t({x} \cond {x}(0))$. Specifically, we can use \eqref{eq:sde_gauss_mean_var} to compute the ground truth score as
\begin{equation}
    \nabla_{{x}}\log p_t({x} \cond {x}(0)) = - \frac{{x}(t) - m_{t}({x})}{v_{t}}.
    \label{eq:score}
\end{equation}
This is analogous to the standard denoising score-matching which also computes the ground truth score based on a clean image and its noisy counterpart~\cite{hyvarinen2005estimation}. Moreover, if we reparameterize ${x}(t) = m_{t}({x}) + \sqrt{v_{t}} \, \epsilon_t$, where $\epsilon_t$ is a standard Gaussian noise $\epsilon_t \sim \mathcal{N}(0, I)$, we can obtain the score directly in terms of the noise by
\begin{equation}
    \nabla_{{x}}\log p_t({x} \cond {x}(0)) = - \frac{\epsilon_t}{\sqrt{v_t}}.
    \label{eq:noise2score}
\end{equation}
Then, we follow the common practice of approximating the noise using a noise network~\cite{ho2020denoising}, i.e. a conditional time-dependent neural network $\tilde{\epsilon}_{\phi}({x}(t), \mu, t)$ which takes both state $x$, condition $\mu$, and time $t$ as input and outputs pure noise. Such a network can be trained with the following objective similar to that used in DDPM~\cite{ho2020denoising}:
\begin{equation}
    L_{\gamma}(\phi) \coloneqq \sum_{i=1}^T \gamma_i \, \mathbb{E} \Big[ \big\lVert \tilde{\epsilon}_\phi({x}_{i}, \mu, i) - \epsilon_{i} \bigr\rVert\Big],
    \label{eq:noise_objective}
\end{equation}
where $\gamma_1, \ldots, \gamma_T$ are positive weights and $\{{x}_i\}_{i=0}^T$ denotes the discretization of the diffusion process. Once trained, we can use the network $\tilde{\epsilon}_{\phi}$ to generate high-quality images by sampling a noisy state ${x}_T$ and iteratively solving the IR-SDE~\eqref{eq:reverse-irsde} with a numerical scheme, such as Euler--Maruyama or Milstein's method~\cite{mil1975approximate}.

\subsection{Maximum Likelihood Learning}

Despite the fact that the objective in~\eqref{eq:noise_objective} offers a simple way to learn the score, we empirically found that the training  often becomes unstable when applied to the complicated degradations encountered in image restoration. We conjecture that this difficulty stems from trying to learn the instantaneous noise at a given time. We therefore propose an alternative maximum likelihood objective, based on the idea of trying to find the optimal trajectory ${x}_{1:T}$ given the high-quality image $x_0$. Note that this objective is not proposed to learn a more accurate score function. Instead, it is used to stabilize training and recover more accurate images. 

Specifically, we want to maximize the likelihood $p({x}_{1:T} \mid {x}_0)$ which can be factorized according to
\begin{equation}
    p({x}_{1:T} \mid {x}_0) = p({x}_{T} \mid {x}_0) \prod_{i=2}^T p({x}_{{i-1}} \mid {x}_{i}, {x}_0),
    \label{eq:joint_objective}
\end{equation}
where $p({x}_T \mid {x}_0)=\mathcal{N}({x}_T; m_{T}({x}_0), v_{T})$ is the low-quality image distribution. Then the reverse transition can be derived from Bayes' rule~\cite{lindholm2022machine}:
\begin{equation}
    p({x}_{{i-1}} \mid {x}_{i}, {x}_0) = \frac{p({x}_{i} \mid {x}_{{i-1}}, {x}_0) p({x}_{{i-1}} \mid {x}_0)}{p({x}_{i} \mid {x}_0)}.
    \label{eq:cond_prob_bayes}
\end{equation}
Since all distributions are Gaussians that can be computed from Proposition \ref{prop:forward_sde_solution}, it is natural to directly find an optimal reverse state that minimizes the negative log-likelihood:
\begin{equation}
\begin{split}
    {x}_{i-1}^{*} = \arg\min_{{x}_{i-1}} \Bigl[ -\log p \bigl({x}_{i-1} \mid {x}_i, {x}_0 \bigr) \Bigr],
    \label{eq:cond_prob_nll}
\end{split}
\end{equation}
where we let ${x}_{i-1}^{*}$ represent the ideal state reversed from ${x}_i$. To simplify the notation, we let $\theta_i^{'} \coloneqq \int_{i-1}^i \theta_t dt$. By solving for the above objective, we have the following:

\begin{proposition}
    Given an initial state ${x}_0$, for any state ${x}_i$ at discrete time $i > 0$, the optimum reversing solution ${x}_{i-1}^{*}$ in (\ref{eq:cond_prob_nll}) for IR-SDE is given by:
    \begin{equation}
    \begin{split}
        {x}_{i-1}^{*} &= \frac{1 - \expp^{-2 \, \bar{\theta}_{i-1}}}{1 - \expp^{-2 \, \bar{\theta}_i}} \expp^{-\theta_i^{'}} ({x}_i - \mu) \\[.6em]
        &\quad+ \frac{1 - \expp^{-2 \, \theta_i^{'}}}{1 - \expp^{-2 \, \bar{\theta}_i}} \expp^{-\bar{\theta}_{i-1}} ({x}_0 - \mu) + \mu.
    \end{split}
    \end{equation}
\end{proposition}
The proof is provided in Appendix~\ref{prf:optimum_trajectory}. Note that we can also use this objective to derive the mean of DDPM\footnote{Please refer to Appendix~\ref{app-sec:mlo_ddpm} for details.}.
Then we choose to optimize the noise network $\tilde{\epsilon}_{\phi}({x}_i, \mu, i)$ to make the IR-SDE reverse as the optimal trajectory, as
\begin{equation}
    {J}_{\gamma}(\phi) \coloneqq \sum_{i=1}^T \gamma_{i} \, \mathbb{E} \Bigl[ \bigl\lVert \underbrace{{x}_{i} - (\diff {x}_{i})_{\tilde{\epsilon}_\phi}}_{\mathrm{reversed} \, {x}_{i-1}} - \, {x}_{i-1}^{*} \bigr\rVert \Bigr],
    \label{eq:nll_objective}
\end{equation}
where $(\diff {x})_{\tilde{\epsilon}_\phi}$ denotes the reverse-time SDE in (\ref{eq:reverse-irsde}) and its score is predicted by the noise network $\tilde{\epsilon}_\phi$. Note that the expectation of the martingale $\int^t_0 \sigma_s \diff \hat{w}(s)$ is zero, implying that we only have to consider the drift part in $(\diff {x})_{\tilde{\epsilon}_\phi}$.

\begin{table}[t]
\vspace{-2.0mm}
\caption{Quantitative comparison between the proposed IR-SDE with other image deraining approaches on the Rain100H test set.}
\label{table:deraining_rain100h}
\vskip 0.05in
\begin{center}
\begin{small}
\begin{sc}
\resizebox{.96\linewidth}{!}{
\begin{tabular}{lcccc}
\toprule
\multirow{2}{*}{Method} &  \multicolumn{2}{c}{Distortion} & \multicolumn{2}{c}{Perceptual}  \\ \cmidrule(lr){2-3} \cmidrule(lr){4-5}
&  PSNR$\uparrow$ & SSIM$\uparrow$ & LPIPS$\downarrow$ & FID$\downarrow$    \\
\midrule
JORDER  & 26.25 & 0.8349 & 0.197 & 94.58   \\
PReNet  & 29.46 & 0.8990 & 0.128 & 52.67   \\
MPRNet  & 30.41 & 0.8906 & 0.158 & 61.59   \\
MAXIM  & 30.81 & 0.9027 & 0.133 & 58.72  \\

CNN-baseline  & 29.12 & 0.8824 & 0.153 & 57.55  \\
IR-SDE  & \textbf{31.65} & \textbf{0.9041} & \textbf{0.047} & \textbf{18.64} \\

\bottomrule
\end{tabular}
}
\end{sc}
\end{small}
\end{center}
\vskip -0.1in
\end{table}

\begin{table}[t]
\caption{Quantitative comparison between the proposed IR-SDE with other image deraining approaches on the Rain100L test set.}
\label{table:deraining_rain100l}
\vskip 0.05in
\begin{center}
\begin{small}
\begin{sc}
\resizebox{.96\linewidth}{!}{
\begin{tabular}{lcccc}
\toprule
\multirow{2}{*}{Method} &  \multicolumn{2}{c}{Distortion} & \multicolumn{2}{c}{Perceptual}  \\ \cmidrule(lr){2-3} \cmidrule(lr){4-5}
&  PSNR$\uparrow$ & SSIM$\uparrow$ & LPIPS$\downarrow$ & FID$\downarrow$  \\
\midrule
JORDER  & 36.61 & 0.9735 & 0.028 & 14.66  \\
PReNet  & 37.48 & 0.9792 & 0.020 & 10.98  \\
MPRNet  & 36.40 & 0.9653 & 0.077 & 26.79  \\
MAXIM  & 38.06 & 0.9770 & 0.048 & 19.06  \\

CNN-baseline  & 33.17 & 0.9583 & 0.068 & 27.32  \\
IR-SDE  & \textbf{38.30} & \textbf{0.9805} & \textbf{0.014} & \textbf{7.94} \\

\bottomrule
\end{tabular}
}
\end{sc}
\end{small}
\end{center}
\vskip -0.1in
\end{table}

\begin{figure}[t]
\begin{center}
\centerline{\includegraphics[width=1.\columnwidth]{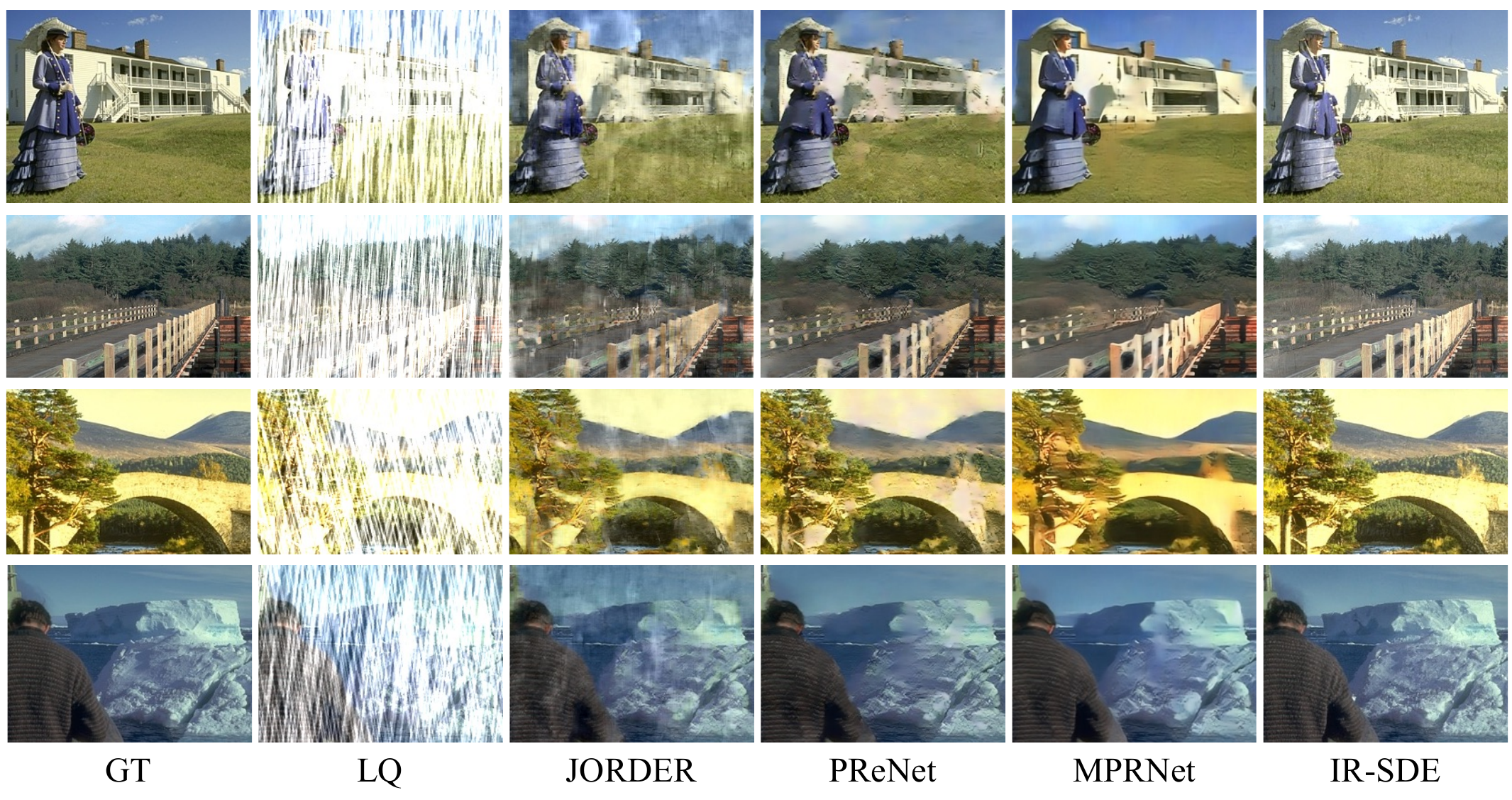}}\vspace{-2.0mm}
\caption{Visual results of our IR-SDE method and other \emph{deraining} approaches on the Rain100H dataset.}
\label{fig:deraining_results}
\end{center}
\vskip -0.4in
\end{figure}

\section{Experiments}
\label{section:experiments}

We experimentally evaluate our proposed IR-SDE method on three popular image restoration tasks: image deraining, deblurring and denoising. 
We compare IR-SDE to the prevailing approaches in their respective fields. The performance of a CNN baseline is also reported in each subsection. The CNN baseline takes a low-quality image as input and directly outputs a high-quality version. It uses the same network architecture as our IR-SDE, but is trained by minimizing the $L_1$ loss between outputs and ground truth images. In addition, we further propose a special SDE and an ordinary differential equation (ODE) to address the Gaussian denoising task. For all tasks, the Learned Perceptual Image Patch Similarity (LPIPS)~\cite{zhang2018unreasonable} and Fr\'{e}chet inception distance (FID)~\cite{heusel2017gans} are reported to measure the perceptual discrepancy and visual effect. The PSNR and SSIM~\cite{wang2004image} are also provided to measure the pixel/structure similarity. Furthermore, we qualitatively illustrate the proposed method on image super-resolution, inpainting, and dehazing tasks. This shows that our method generalizes well to various image restoration problems, and \emph{the only change required for each task was to change the dataset}. Implementation details are provided in Appendix \ref{app-sec:implementation}. For each of the six image restoration tasks, additional qualitative results are also found in Appendix~\ref{app-sec:results}.

\subsection{Image Deraining}

We evaluate IR-SDE on two synthetic raining datasets: Rain100H~\cite{yang2017deep} and Rain100L~\cite{yang2017deep}. The former has 1\thinspace800 pairs of images with/without rain for training, and 100 pairs for testing. The latter has 200 pairs for training and 100 pairs for testing. In this task, we report PSNR and SSIM scores on the Y channel (YCbCr space) similar to existing deraining methods~\cite{ren2019progressive,zamir2021multi}. Moreover, we compare our methods with several state-of-the-art deraining approaches such as JORDER~\cite{yang2019joint}, PReNet~\cite{ren2019progressive}, MPRNet~\cite{zamir2021multi}, and MAXIM~\cite{tu2022maxim}. 
Note that achieving state-of-the-art performance on a specific task is not the main focus of this paper. Similar to other diffusion approaches, we will place more attention on the perceptual scores. 

The quantitative comparisons on the two raining datasets are shown in Tables~\ref{table:deraining_rain100h} and~\ref{table:deraining_rain100l}.  The proposed IR-SDE achieves the best performance in all metrics. In particular, the perceptual scores (LPIPS and FID) of the IR-SDE are markedly better than those of the other approaches. Based on these scores and the visual comparison in Figure~\ref{fig:deraining_results}, we conclude that IR-SDE clearly produces the most realistic and high-fidelity results. Moreover, the CNN-baseline model only outperforms JORDER. Our method significantly improves its performance without changing the network structure, which further illustrates the superiority of the proposed method.

\begin{table}[t]
\vspace{-2.0mm}
\caption{Quantitative comparison between the proposed IR-SDE with other image deblurring approaches on the GoPro test set.}
\label{table:deblurring}
\vskip 0.05in
\begin{center}
\begin{small}
\begin{sc}
\resizebox{.96\linewidth}{!}{
\begin{tabular}{lcccc}
\toprule
\multirow{2}{*}{Method} &  \multicolumn{2}{c}{Distortion} & \multicolumn{2}{c}{Perceptual}  \\ \cmidrule(lr){2-3} \cmidrule(lr){4-5}
&  PSNR$\uparrow$ & SSIM$\uparrow$ & LPIPS$\downarrow$ & FID$\downarrow$    \\
\midrule
DeepDeblur  & 29.08 & 0.9135 & 0.135 & 15.14   \\
DeblurGAN  & 28.70 & 0.8580 & 0.178 & 27.02   \\
DeblurGAN-v2  & 29.55 & 0.9340 & 0.117 & 13.40   \\
DBGAN  & 31.18 & 0.9164 & 0.112 & 12.65   \\
MAXIM   & \textbf{32.86} & \textbf{0.9403} & 0.089 & 11.57   \\
CNN-baseline  & 28.87 & 0.8469 & 0.225 & 23.09  \\
IR-SDE  & 30.70 & 0.9010 & \textbf{0.064} & \textbf{6.32}  \\

\bottomrule
\end{tabular}
}
\end{sc}
\end{small}
\end{center}
\vskip -0.1in
\end{table}

\begin{figure}[t]
\begin{center}
\centerline{\includegraphics[width=1.\columnwidth]{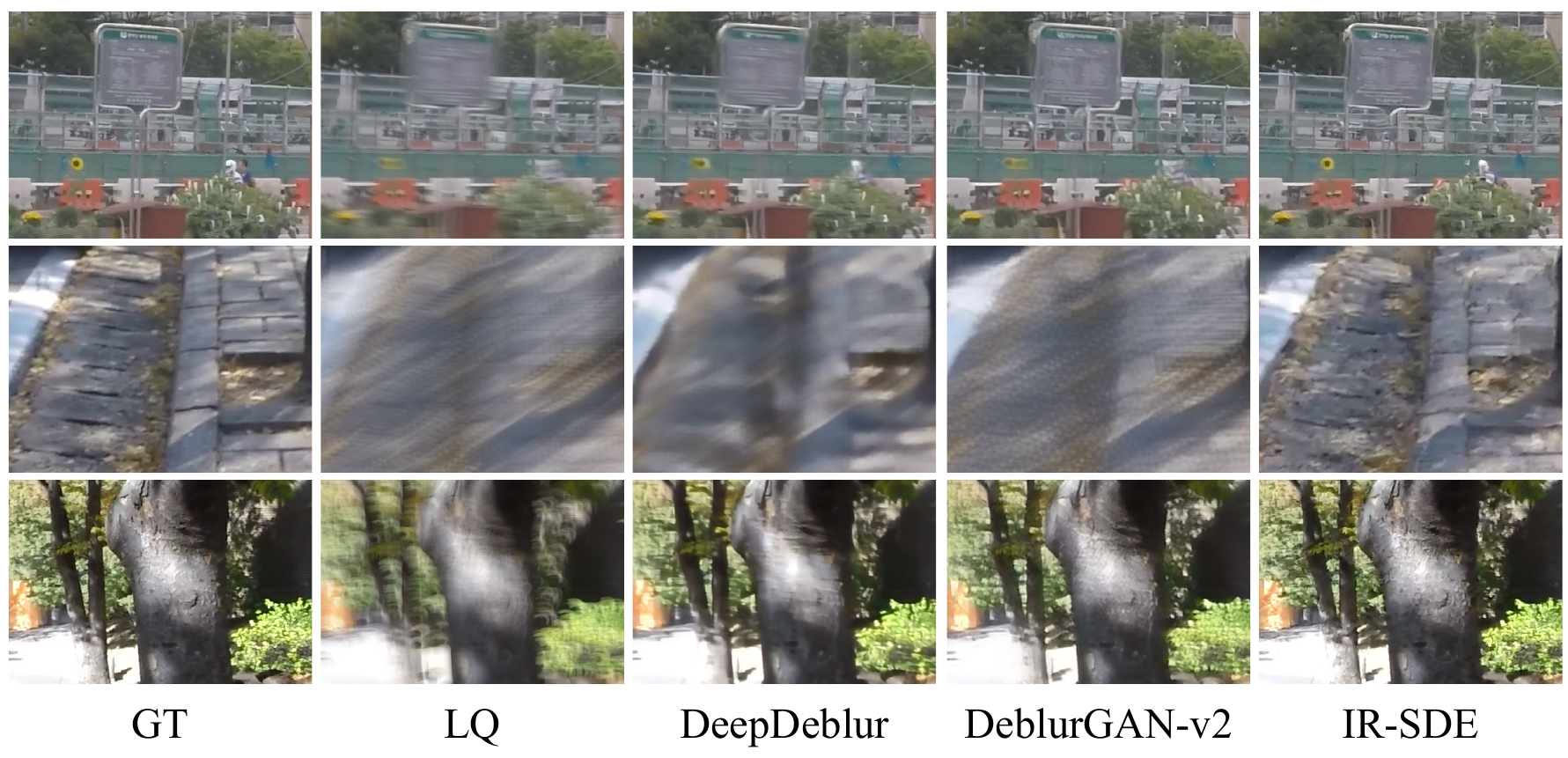}}\vspace{-2.0mm}
\caption{Visual results of our IR-SDE method compared to other \emph{deblurring} approaches on the GoPro dataset.}
\label{fig:deblurring_results}
\end{center}
\vskip -0.2in
\end{figure}

\subsection{Image Deblurring}

We evaluate the deblurring performance of IR-SDE on the public GoPro dataset~\cite{nah2017deep} which contains 2\thinspace103 image pairs for training and 1\thinspace111 image pairs for testing. Note that the blurry images in GoPro are collected by averaging multiple sharp images captured by a high-speed video camera. Compared with other synthetic blurry images from blur kernels, the GoPro dataset contains more realistic blur and is much more complex. 

Table~\ref{table:deblurring} summarizes the quantitative results of image deblurring. For comparison, we report four milestone deblurring approaches: DeepDeblur~\cite{nah2017deep}, DeblurGAN~\cite{kupyn2018deblurgan}, DeblurGAN-v2~\cite{kupyn2019deblurgan}, DBGAN~\cite{zhang2020deblurring}, and MAXIM~\cite{tu2022maxim}. Our method surpasses DeblurGAN-v2 by 1.15 dB in terms of PSNR and achieves the best perceptual performance overall. This indicates that the sharp images produced by IR-SDE look more realistic than other GAN-based methods and are still consistent with the ground truths. Moreover, our method significantly improves the CNN-baseline without changing its network structure, which also illustrates the superiority of our method. The visual comparison in Figure~\ref{fig:deblurring_results} shows that our method is able to handle difficult blurring cases and produces mostly clear and visually appealing results.

\begin{figure}[t]
\begin{center}
\centerline{\includegraphics[width=1.\columnwidth]{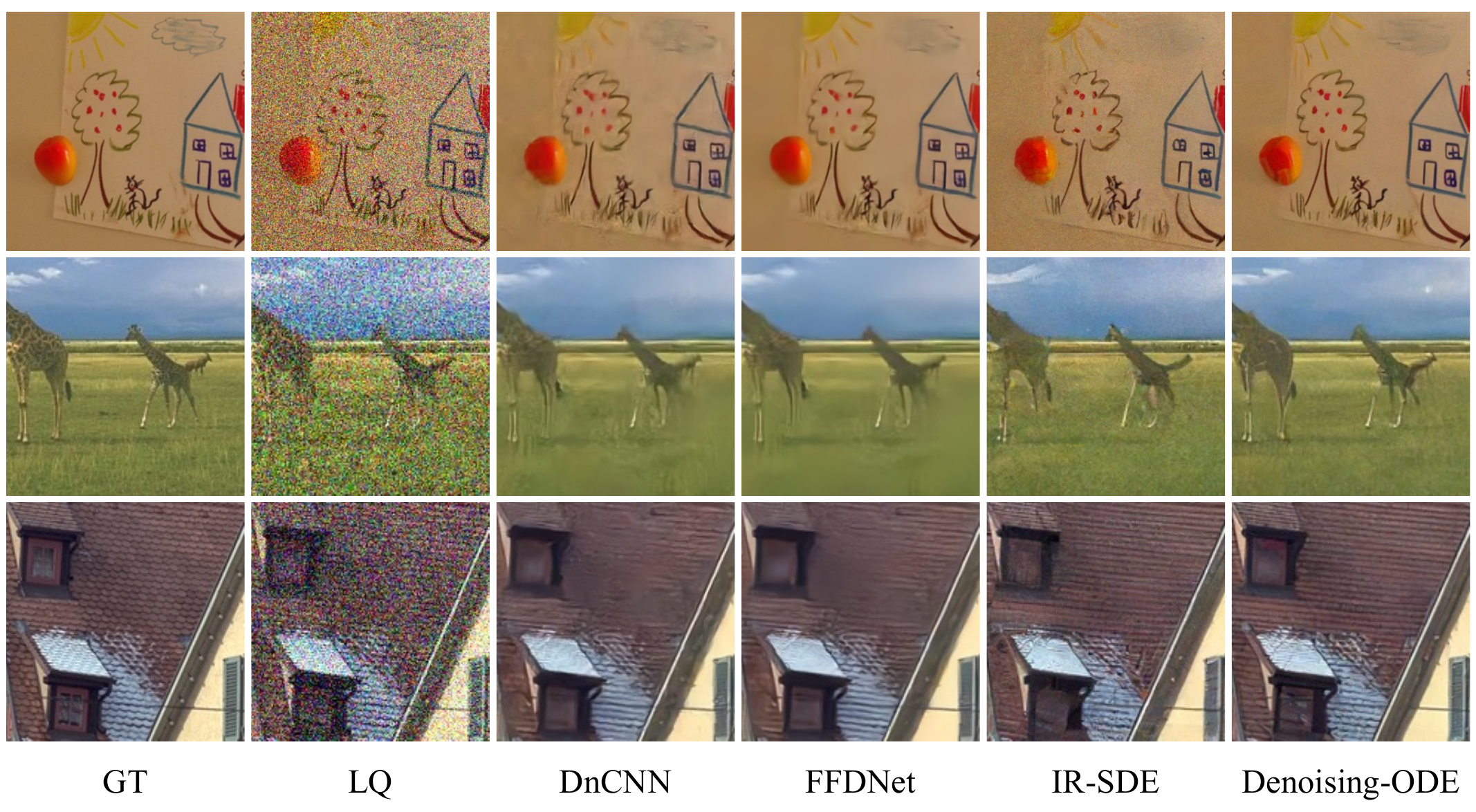}}\vspace{-2.0mm}
\caption{Visual results of our methods with other \emph{denoising} approaches. The total timesteps of IR-SDE is fixed to 100, while the Denoising ODE only requires $\textit{22}$ steps to recover the clean image.}
\label{fig:denoising_results}
\end{center}
\vskip -0.2in
\end{figure}

\begin{table*}[t]
\caption{Denoising results on different test sets with noise level $\sigma=25$. Note that the total steps of IR-SDE is 100, while the Denoising SDE/ODE only require $\textit{22}$ steps to recover the clean image. We provide more details and results in Appendix~\ref{app-sec:denoising_sde} and \ref{app-sec:results}, respectively.}\vspace{-2.5mm}
\label{table:denoising_results}
\vskip 0.05in
\begin{center}
\begin{small}
\begin{sc}
\resizebox{1.\linewidth}{!}{
\begin{tabular}{lcccccccccccc}
\toprule
\multirow{2}{*}{Method} &  \multicolumn{4}{c}{McMaster} & \multicolumn{4}{c}{Kodak24} & \multicolumn{4}{c}{CBSD68} \\ \cmidrule(lr){2-5} \cmidrule(lr){6-9} \cmidrule(lr){10-13}
&  PSNR$\uparrow$ & SSIM$\uparrow$ & LPIPS$\downarrow$ & FID$\downarrow$ &  PSNR$\uparrow$ & SSIM$\uparrow$ & LPIPS$\downarrow$ & FID$\downarrow$ &  PSNR$\uparrow$ & SSIM$\uparrow$ & LPIPS$\downarrow$ & FID$\downarrow$  \\
\midrule
DnCNN  & 31.52 & 0.8692 & 0.101 & 59.16 & 32.02 & 0.8763 & 0.129 & 41.96 & 31.24 & 0.8830 & 0.109 & 43.51 \\
FFDNet  & 32.36 & \textbf{0.8861} & 0.103 & 63.84 & 32.13 & \textbf{0.8779} & 0.140 & 44.57 & \textbf{31.22} & \textbf{0.8821} & 0.121 & 49.64 \\
CNN-baseline  & 31.79 & 0.8697 & 0.122 & 66.47 & 32.73 & 0.8666 & 0.161 & 45.81 & 30.74 & 0.8661 & 0.162 & 56.64 \\

\midrule

IR-SDE  & 29.48 & 0.8052 & 0.071 & 44.77 & 28.99 & 0.7772 & 0.106 & 35.19 & 28.09 & 0.7866 & 0.101 & 36.49 \\
Denoising-SDE  & 28.98 & 0.7512 & 0.088 & 45.84 & 28.55 & 0.7247 & 0.130 & 36.18 & 27.65 & 0.7457 & 0.131 & 39.25 \\
Denoising-ODE  & \textbf{32.39} & 0.8791 & \textbf{0.055} & \textbf{34.66} & \textbf{32.14} & 0.8739 & \textbf{0.078} & \textbf{21.47} & 31.14 & 0.8777 & \textbf{0.074} & \textbf{28.71} \\

\bottomrule
\end{tabular}
}
\end{sc}
\end{small}
\end{center}
\vskip -0.1in
\end{table*}

\begin{table*}[t]
\begin{center}
\caption{Comparison of our method with DDRM~\cite{kawar2022denoising} on Gaussian image denoising, super-resolution, and face inpainting. We use the CBSD68, DIV2K, and CelebA-HQ datasets for task evaluations, respectively. Note that DDRM requires that the degradation parameters are known and can be composed by SVD. Moreover, all images are center cropped with size $256 \times 256$. }\vspace{-2.5mm}
\label{table:ddrm}
\vskip 0.05in

\begin{small}
\begin{sc}
\resizebox{1.\linewidth}{!}{
\begin{tabular}{lcccccccccccc}
\toprule
\multirow{2}{*}{Method} &  \multicolumn{4}{c}{Image Denoising} & \multicolumn{4}{c}{Super-Resolution} & \multicolumn{4}{c}{Face Inpainting} \\ \cmidrule(lr){2-5} \cmidrule(lr){6-9} \cmidrule(lr){10-13}
&  PSNR$\uparrow$ & SSIM$\uparrow$ & LPIPS$\downarrow$ & FID$\downarrow$ &  PSNR$\uparrow$ & SSIM$\uparrow$ & LPIPS$\downarrow$ & FID$\downarrow$ &  PSNR$\uparrow$ & SSIM$\uparrow$ & LPIPS$\downarrow$ & FID$\downarrow$  \\
\midrule
DDRM  & 29.57 & 0.8484 & 0.135 & 58.99 & 24.35 & 0.5927 & 0.364 & 78.71 & 27.16 & 0.8893 & 0.089 & 37.02 \\
CNN-baseline  & 29.69 & 0.8529 & 0.170 & 59.99 & \textbf{26.64} & \textbf{0.6729} & 0.389 & 133.95 & \textbf{29.22} & \textbf{0.9218} & 0.065 & 38.35 \\
Ours  & \textbf{31.01} & \textbf{0.8746} & \textbf{0.069} & \textbf{29.72} & 25.90 & 0.6570 & \textbf{0.231} & \textbf{45.36} & 28.37 & 0.9166 & \textbf{0.046} & \textbf{25.13} \\

\bottomrule
\end{tabular}
}
\end{sc}
\end{small}
\end{center}
\end{table*}

\subsection{Gaussian Image Denoising}

Recall that the Wiener process in the SDE is a Gaussian process. Hence, we introduce a \textit{Denoising-SDE} -- which is a special case of the IR-SDE in \eqref{equ:ou} and \eqref{eq:reverse-irsde} -- such that we can carry out the denoising computations with fewer time steps, by setting the clean image as the mean $\mu = {x}_0$ for all times~$t$.
Thus we can regard any noisy image as an intermediate state and directly reverse it to a clean image. Moreover, since there is only Gaussian noise on the clean image, it is reasonable to derive a denoising ordinary differential equation (ODE) that shares the same marginal probability as the SDE~\cite{song2021score} but can perform denoising without introducing additional noise from a Wiener process. This \textit{Denoising-ODE} is given by,
\begin{equation}
    \diff {x} = \Bigl[\theta_t \, (\mu - {x}) - \frac{1}{2} \, \sigma_t^2 \, \nabla_{{x}}\log p_t({x}) \Bigr]\diff t.
    \label{eq:reverse-irode}
\end{equation} 
Theoretically, we can use \eqref{eq:reverse-irode} to solve the Gaussian denoising problem deterministically. The main difference between Denoising SDE and ODE is the stochastic term (i.e., Wiener process). In Appendix~\ref{app-sec:denoising_sde}, we provide a detailed deduction for the Denoising SDE/ODE and show that we can derive an appropriate denoising step to improve the sample efficiency.

To evaluate the image denoising performance, we train our models on 8\thinspace294 high-quality images collected from the DIV2K~\cite{agustsson2017ntire}, Flickr2K~\cite{timofte2017ntire}, BSD500~\cite{arbelaez2010contour}, and Waterloo Exploration datasets~\cite{ma2016waterloo}. 
Then all models are evaluated on the McMaster~\cite{zhang2011color}, Kodak24~\cite{franzen1999kodak}, and CBSD68~\cite{martin2001database} datasets. To show that our methods are in line with the state-of-the-art, we compare to the methods of~\cite{zhang2017beyond} and~\cite{zhang2018ffdnet}, which we call DnCNN and FFDNet, respectively.

The numerical results for the three test datasets are reported in Table~\ref{table:denoising_results}. The IR-SDE has a high perceptual performance but its fidelity scores (i.e., PSNR and SSIM) are worse than other CNN-based methods, and the same goes for Denoising-SDE. 
The reason may be that the diffusion process is not identifiable from the Gaussian noises, because the Denoising-ODE, which does not have the stochastic term, has significantly better PSNR on all datasets. The visual comparisons are shown in Figure~\ref{fig:denoising_results}. One can see that CNN-based methods very often produce over-smoothed images. Although IR-SDE and Denoising-ODE both generate realistic results, those of the Denoising-ODE are less noisy. We also compare Denoising-ODE with the recent diffusion method DDRM~\cite{kawar2022denoising} on cropped images in Table~\ref{table:ddrm}, achieving superior performance across all metrics.

\begin{figure}[t]
\begin{center}
\centerline{\includegraphics[width=1.\columnwidth]{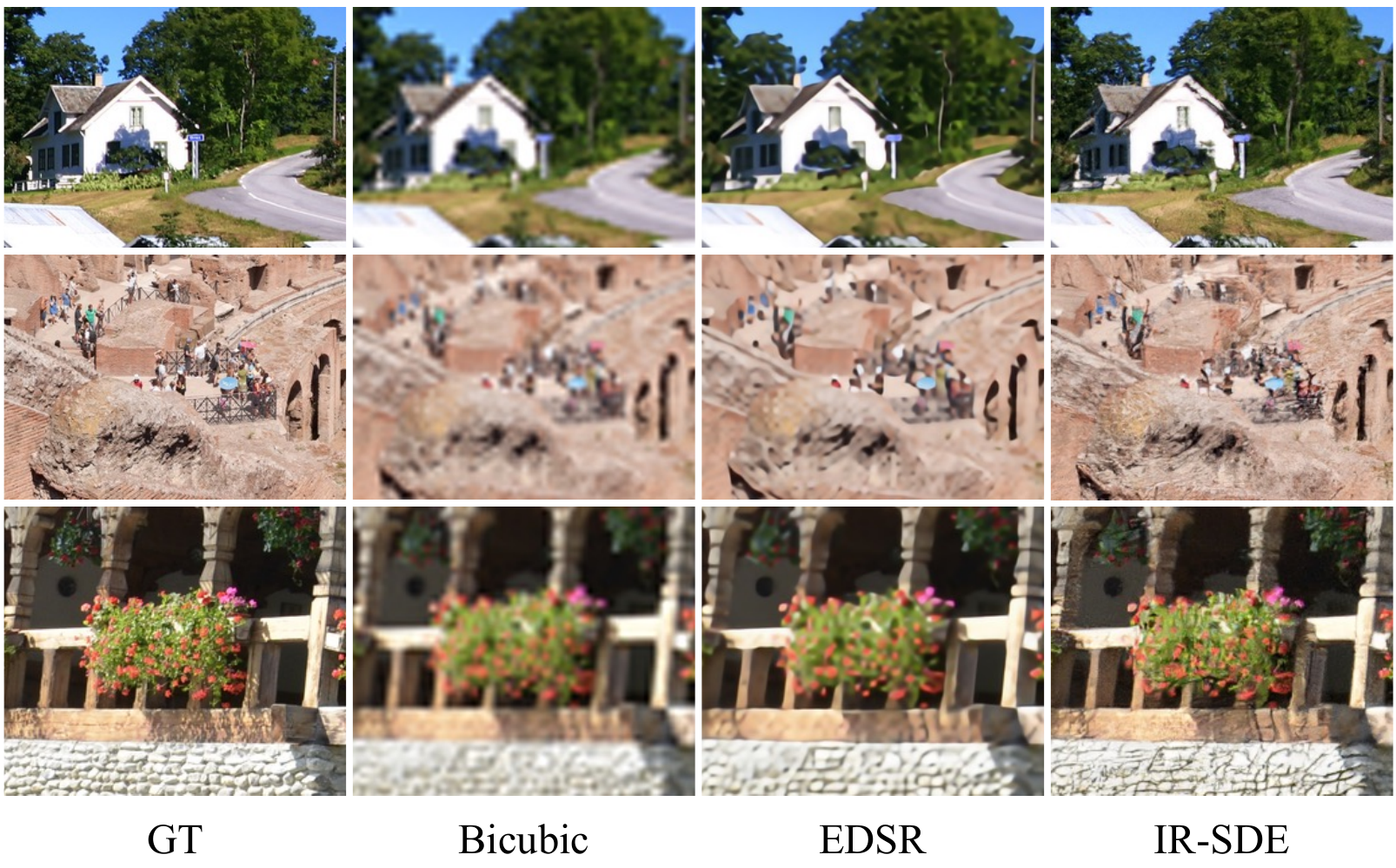}}\vspace{-2.0mm}
\caption{Visual results of our IR-SDE method with EDSR on the DIV2K validation dataset for \emph{super-resolution}. The LQ images are bicubicly upsampled to have the same size as GT images.}
\label{fig:sr_results}
\end{center}
\vskip -0.2in
\end{figure}

\begin{figure*}[ht]
\begin{center}
\centerline{\includegraphics[width=1.\linewidth]{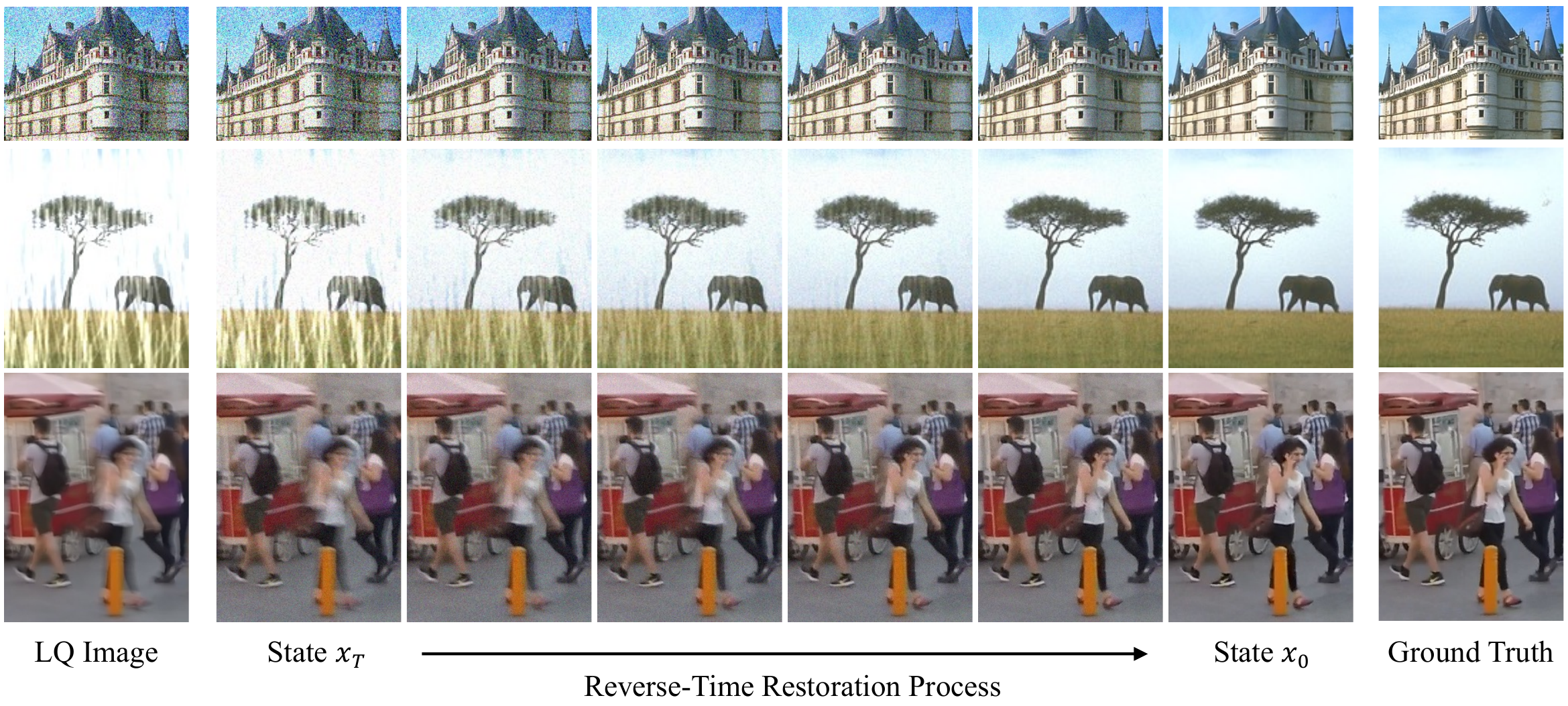}}\vspace{-2.0mm}
\caption{Illustration of the visual process of the reverse-time image restoration. The top row is the example of denoising ODE, in which the state ${x}(T)$ is exactly the LQ image. Middle and bottom rows are examples of IR-SDE for deraining and deblurring, respectively.}
\label{fig:show_reverse_time_restoration}
\end{center}
\vskip -0.2in
\end{figure*}

\begin{figure}[t]
\begin{center}
\centerline{\includegraphics[width=1.\columnwidth]{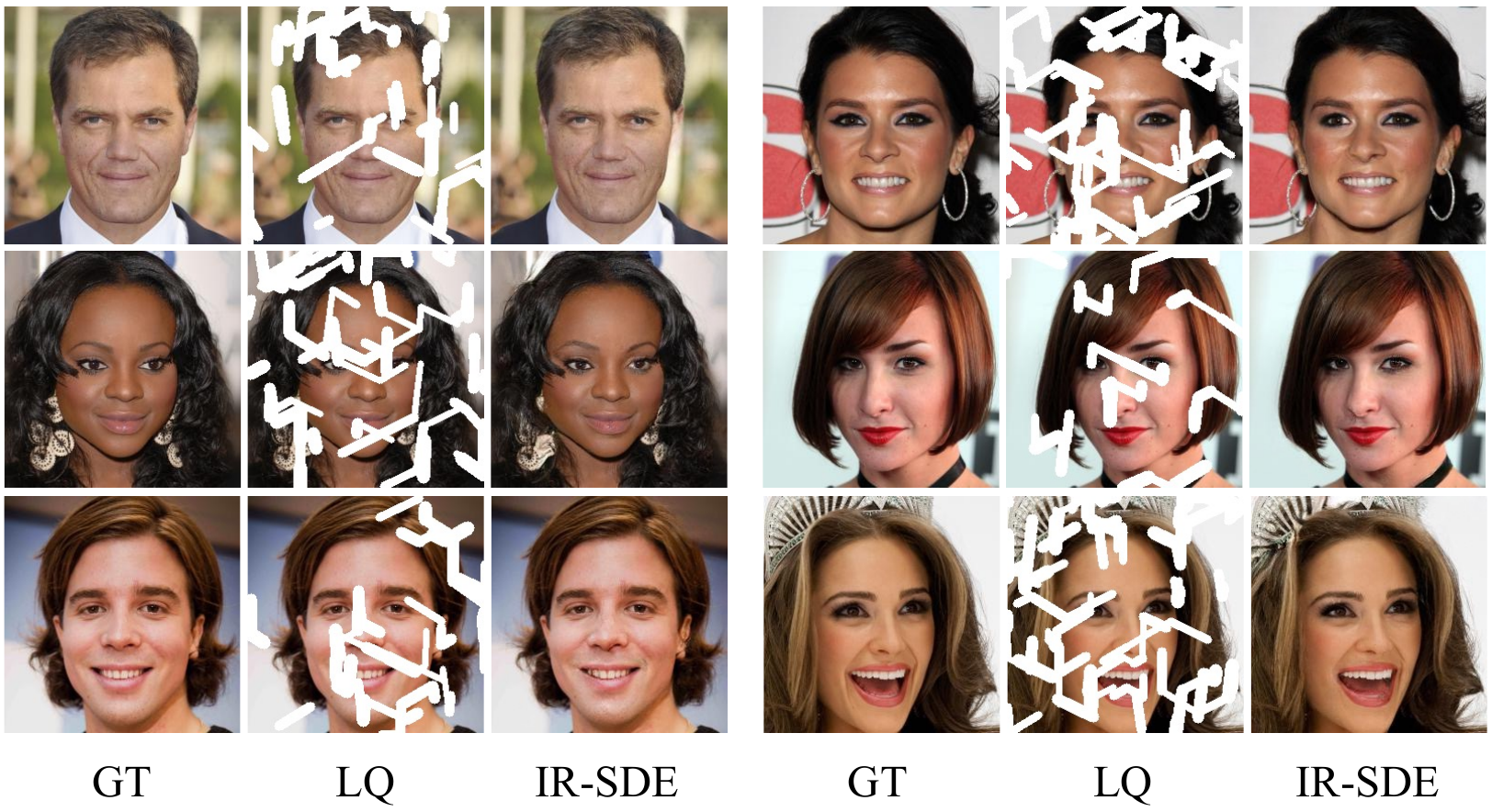}}\vspace{-2.0mm}
\caption{Visual results of \emph{inpainting} on the CelebA-HQ dataset.}
\label{fig:inpainting_results}
\end{center}
\vskip -0.2in
\end{figure}

\begin{figure}[t]
\begin{center}
\centerline{\includegraphics[width=1.\columnwidth]{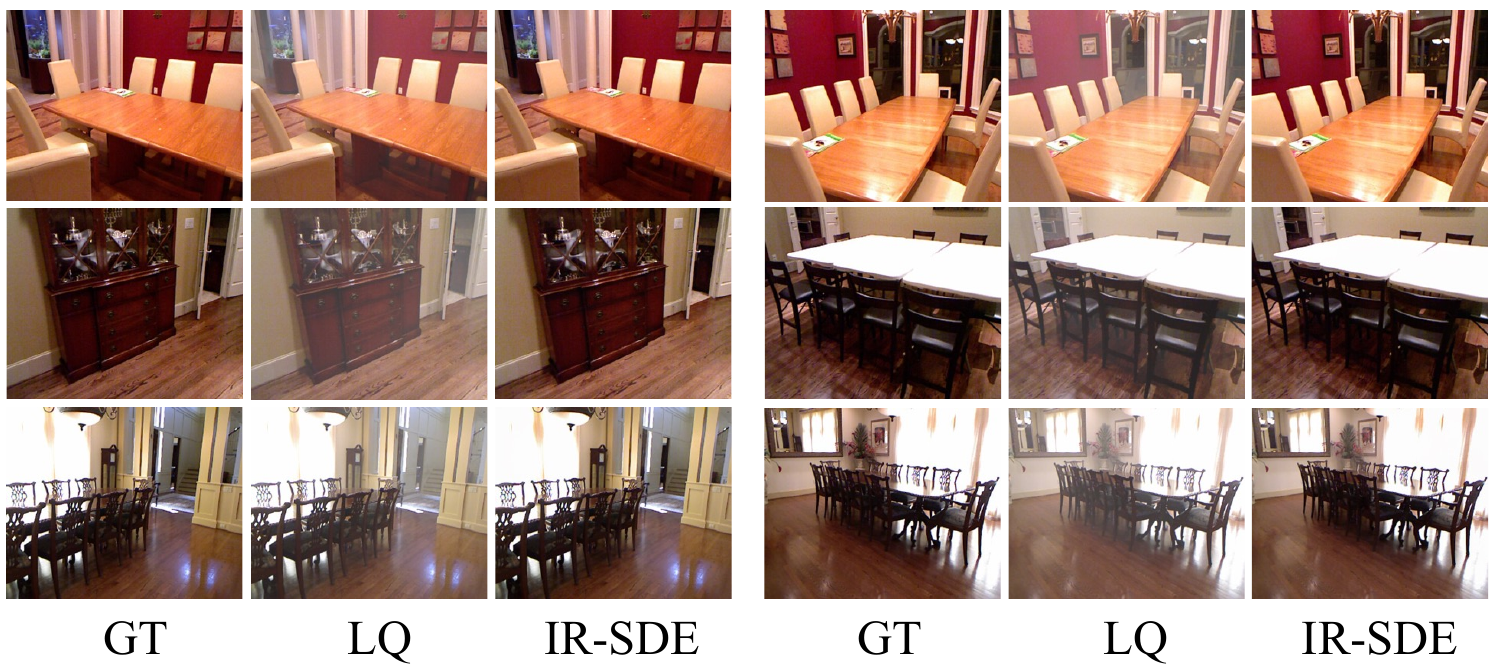}}\vspace{-2.0mm}
\caption{Visual results of \emph{dehazing} on the SOTS indoor dataset.}
\label{fig:dehazing_results}
\end{center}
\vskip -0.2in
\end{figure}

\subsection{Qualitative Experiments}
\label{secton:experiments:qualitative}

In this section, we further demonstrate the general applicability of our proposed IR-SDE method by performing qualitative experiments on image super-resolution, inpainting, and dehazing. The training settings for these experiments are the same as those of the previous sections. For super-resolution and inpainting, we also compare our quantitative results with DDRM~\cite{kawar2022denoising} to show the superiority of our method.

\parsection{Super-Resolution}
We first experiment on single image super-resolution, which is a fundamental and challenging task in computer vision. Our IR-SDE is trained and evaluated on the DIV2K~\cite{agustsson2017ntire} dataset. As an additional preprocessing step, all the low-resolution images are bicubicly re-scaled to be of the same size as the corresponding high-resolution images. Figure~\ref{fig:sr_results} shows the qualitative results on the DIV2K validation dataset. Compared to the $L_2$ trained EDSR~\cite{lim2017enhanced} model, our IR-SDE is able to restore images that have rich details and are visually clear and realistic. Here we also provide the quantitative comparison with another diffusion-based model DDRM~\cite{kawar2022denoising} in Table~\ref{table:ddrm}.

\parsection{Face Inpainting}
Inpainting is the task of filling new content to missing regions of an image. We select the CelebA-HQ~\cite{karras2018progressive} dataset to train and test the IR-SDE on this task. Here we set the mask to be unknown. The inpainted regions must harmonize with the rest such that the overall face is semantically reasonable and has a natural appearance. Visual examples of face inpainting are illustrated in Figure~\ref{fig:inpainting_results}. As can be seen, the proposed IR-SDE demonstrates a strong generative capability in restoring masked areas while it at the same time maintains consistency with the original image. Moreover, the quantitative comparison with DDRM~\cite{kawar2022denoising} is shown in Table~\ref{table:ddrm}.

\parsection{Dehazing}
Image dehazing is often an important prerequisite for improving the robustness of other high-level vision tasks. Note that DDRM requires that the degradation parameters are known and can be composed by SVD, and therefore not can be applied to dehazing. In contrast, our method is flexible to deal with all kinds of tasks. We train the IR-SDE on the RESIDE~\cite{li2018benchmarking} Indoor Training Set (ITS) and test it on the Synthetic Objective Testing Set (SOTS). As shown in Figure~\ref{fig:dehazing_results}, our IR-SDE successfully restores haze-free indoor scenes from the low-quality and low-contrast inputs. The quantitative results are shown in Appendix~\ref{app-sec:results}.

\section{Discussion and Analysis}
\label{section:discussion}

In this section, we first give an in-depth analysis of the reverse-time restoration process of the IR-SDE, and then study two important components (maximum likelihood objective and theta schedule) and the limitations in more detail.

\begin{figure}[b]
\begin{center}
\centerline{\includegraphics[width=.95\columnwidth]{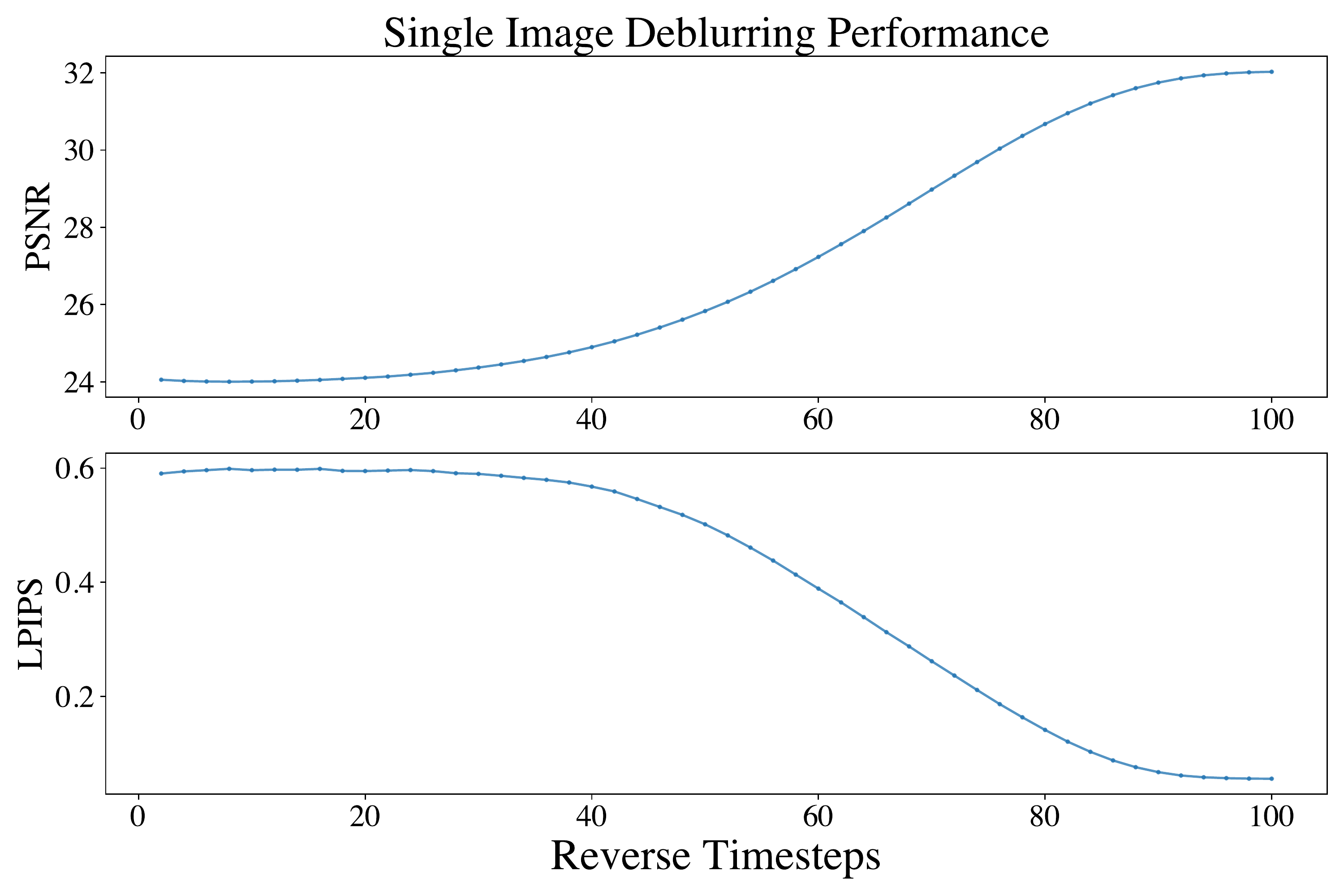}}\vspace{-2.0mm}
\caption{Performance curves of the reverse-time deblurring process. Curves are computed from the deblurring case in Figure~\ref{fig:show_reverse_time_restoration}.
}
\label{fig:deblurring_test_curve}
\end{center}
\vskip -0.2in
\end{figure}

\subsection{Reverse-Time Restoration Process}

For IR-SDE, the terminal state ${x}_T$ is generally obtained by adding noise to the degraded low-quality image. To restore a high-quality image, both the degradation and the noise thus have to be gradually removed. But how are these two different corruptions handled in the reverse-time process?

To analyze this we provide a few concrete restoration examples in Figure~\ref{fig:show_reverse_time_restoration}. Note that the top row of Figure~\ref{fig:show_reverse_time_restoration} shows the denoising case by Denoising-ODE, where the noisy image is considered to be an intermediate state and the only aim is to gradually remove the Gaussian noise to recover the clean image. For other image restoration cases, we find that the IR-SDE tends to assign a higher priority to handle the original degradation and only performs Gaussian denoising in the last few steps. As illustrated for the image deraining and deblurring cases in Figure~\ref{fig:show_reverse_time_restoration}, most of the degradation (rain and blur) has been removed already in the middle timesteps.

In addition, we show the performance curves of the IR-SDE (with cosine schedule) when it comes to deblurring a single image in Figure~\ref{fig:deblurring_test_curve}. As can be seen, the deblurring performance (in terms of PSNR and LPIPS) increases after running 20 steps and then converges in the last few steps.

\begin{figure}[t]
\begin{center}
\centerline{\includegraphics[width=1.\columnwidth]{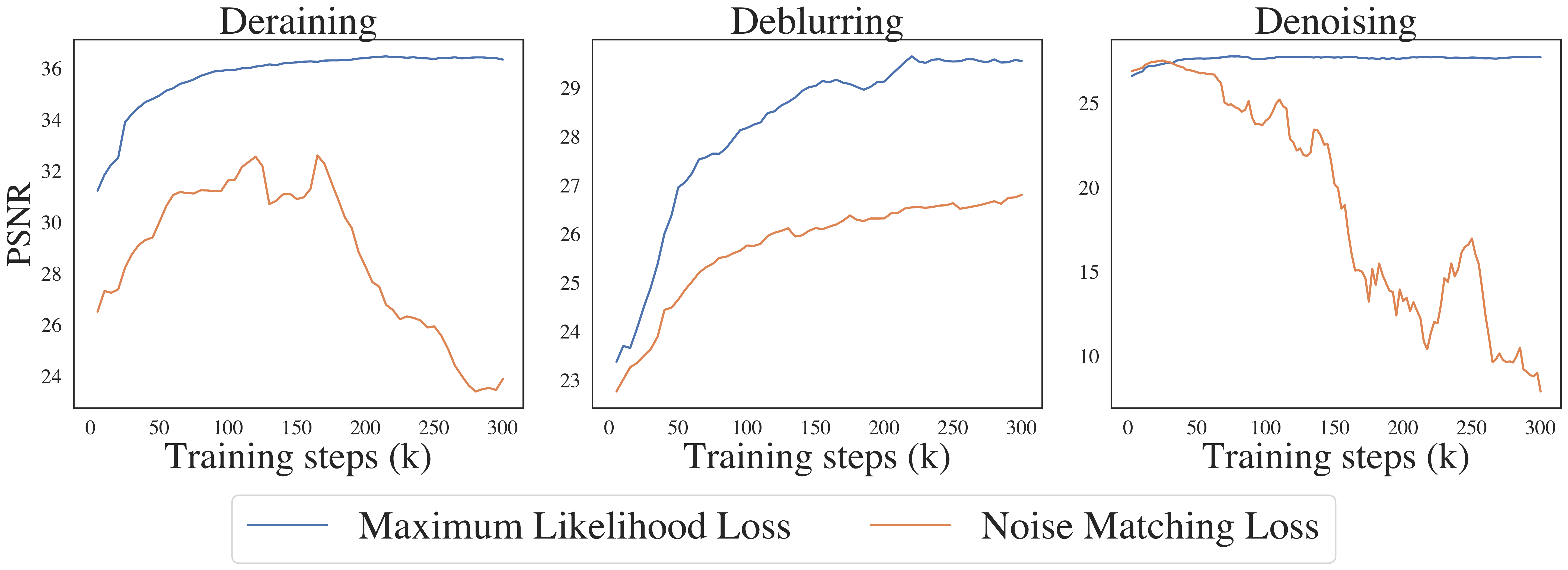}}\vspace{-2.0mm}
\caption{Training curves of the IR-SDE on various tasks. Our proposed maximum likelihood-based loss function stabilizes the training and improves the restoration performance.}
\label{fig:loss_curves}
\end{center}
\vskip -0.2in
\end{figure}

\subsection{Maximum Likelihood Objective}
\label{sec:discuss_objective}
A key improvement of our IR-SDE method compared to other diffusion models, which directly learn the noise/score, is that we learn an optimal reverse-time trajectory from ${x}_T$ to ${x}_0$ based on the maximum likelihood objective in~\eqref{eq:nll_objective}. Here we show that this objective results in more stable training, which in turn improves the restoration performance, as illustrated in Figure~\ref{fig:loss_curves}. The PSNR when training with a noise-matching objective fluctuates and even deteriorates over time in the deraining and denoising tasks. While the training still works for deblurring, the performance is clearly inferior to the proposed maximum likelihood objective.

\subsection{Time-Varying Theta Schedules}
\label{section:discussion:theta}

\begin{figure}[t]
\begin{center}
\centerline{\includegraphics[width=.85\columnwidth]{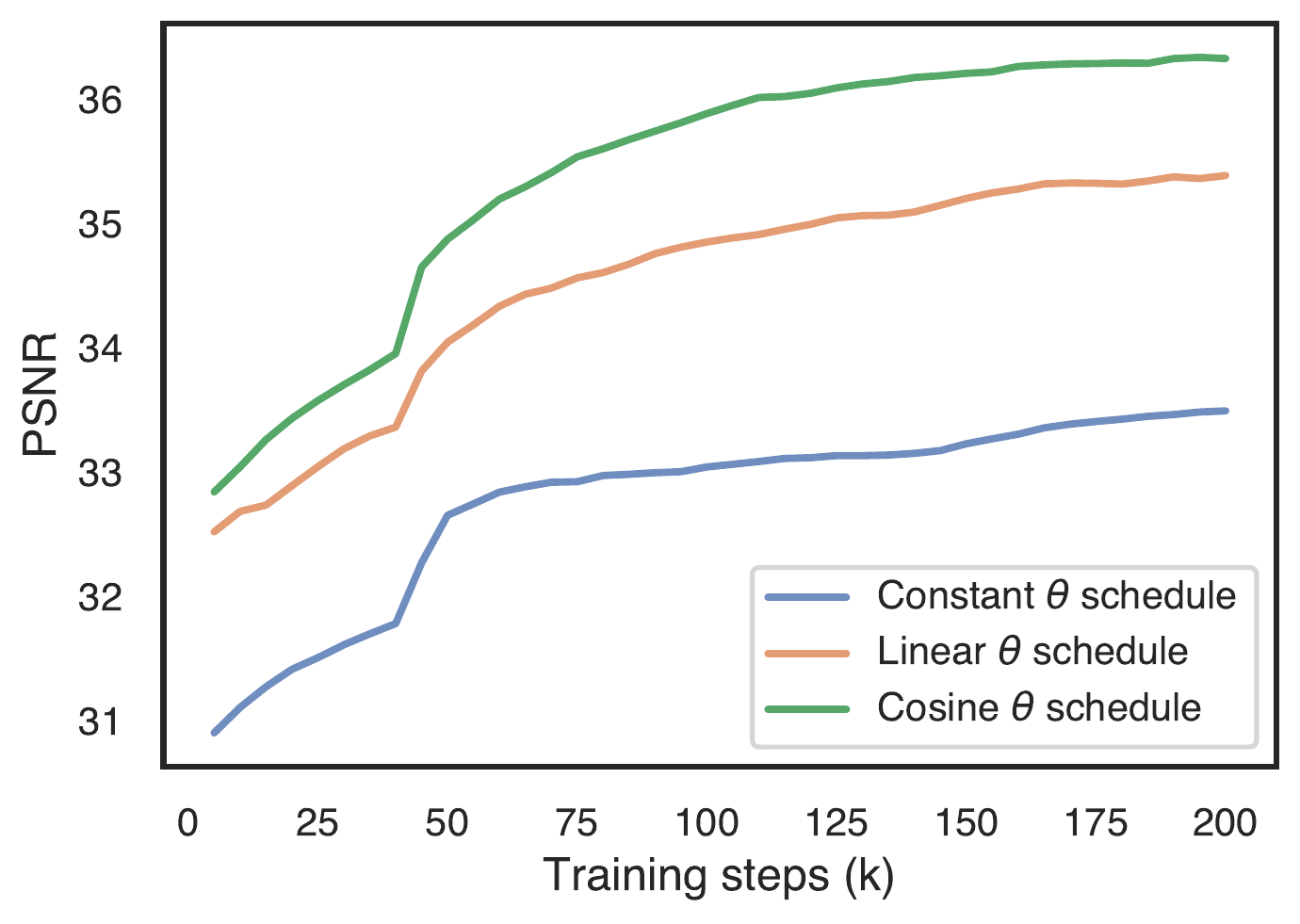}}\vspace{-2.0mm}
\caption{Training curves of different $\theta$ schedules on the image deraining task. Note that when $\theta$ is constant, the forward IR-SDE~\eqref{equ:ou} simplifies to the OU process.}
\label{fig:theta_schedule_curves}
\end{center}
\vskip -0.2in
\end{figure}

It is notable that our IR-SDE has two time-varying parameters $\theta_t$ and $\sigma_t$, which we set to be constrained by the stationary variance of ${x}_t$ as $\sigma_t^2 \, / \, \theta_t = 2 \, \lambda^2$ for all timesteps. Since $\lambda$ is fixed as the noise level applied to the LQ image, we can simply adjust $\theta$ to construct different noise schedules in IR-SDE. As shown in Figure~\ref{fig:theta_schedule_curves}, we explore three different schedules for how to vary $\theta$: constant, linear, and cosine (see Appendix~\ref{app-sec:implementation} for details). When $\theta$ is constant, the IR-SDE simplifies to the Ornstein--Uhlenbeck (OU) process~\cite{gillespie1996exact} which is widely used to solve mean-reverting problems. The linear/cosine schedules are widely-used in existing diffusion probabilistic models~\cite{ho2020denoising,nichol2021improved}. We use their flipped version for $\theta_t$ such that the diffusion coefficient $\sigma_t$ smoothly changes to a maximum value as $t \to \infty$. It is observed that all schedules work well for the deraining task, and that the cosine schedule performs significantly better than others.

\subsection{Limitations and Future Works}
\label{sec:limitations}
\begin{figure}[t]
\begin{center}
\centerline{\includegraphics[width=1.\columnwidth]{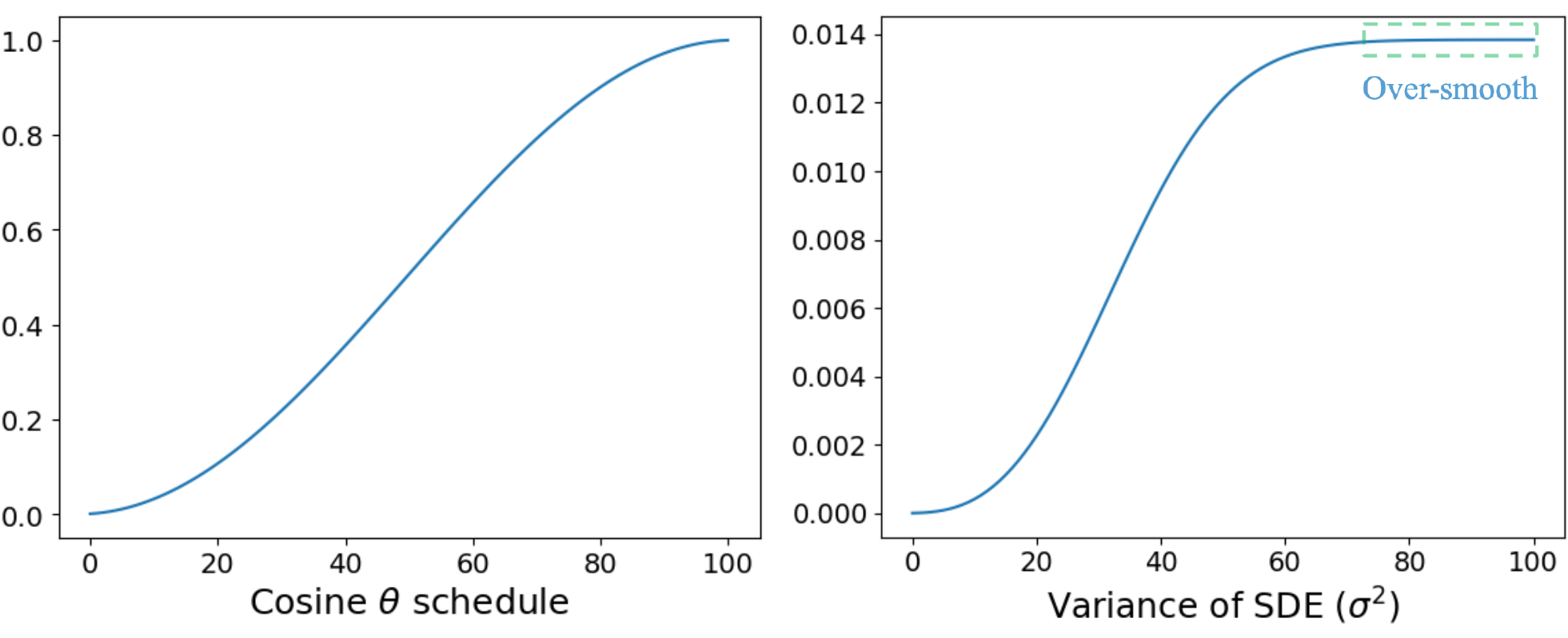}}\vspace{-2.0mm}
\caption{Cosine $\theta$ schedule and the variance of the forward SDE. The variance changes over-smoothly in the last few steps. }
\label{fig:cosine_variance}
\end{center}
\vskip -0.2in
\end{figure}

We have shown the usefulness of our method on various image restoration tasks. However, it is also important to acknowledge one potential limitation: the exponential term in \eqref{eq:sde_gauss_mean_var} for $v_t$ leads to an overly smooth variance change in the last few steps (see Figure \ref{fig:cosine_variance}). In that area, the neighboring states (${x}_i$, ${x}_{i-1}$) have quite similar appearances thus making learning difficult, especially when the maximum likelihood loss (which optimizes the difference between states) is used. In our future work, we will explore alternative theta schedules to alleviate this problem.

Moreover, it is worth noting that we can generalize the choice of the SDE, hence the conditional score, by using Tweedie's formula, see \citet[][Table 1]{Kim2021Tweedie} and~\citet{Kim2022Tweedie}. As an example, if we choose the SDE to be a geometric Brownian motion, then the score in Equation~\eqref{eq:score} corresponds to that of an exponential distribution.
\looseness=-1

\section{Related Work}

Image restoration is an active research topic within computer vision \citep{zhang2017image,zhang2017learning,wang2022uformer,xiao2022stochastic}. The most common approach is to train some type of deep learning model to solve image restoration tasks in a supervised manner~\citep{zamir2021multi}. Various CNN-based architectures have been proposed~\citep{zamir2021multi,chen2022simple}, and recently the use of transformers has also been extensively explored~\citep{liang2021swinir,zamir2022restormer,luo2022bsrt}. These methods all entail training a neural network to directly predict high-quality images from given low-quality ones. In contrast, our proposed IR-SDE approach gradually restores a given low-quality image by simulating the reverse-time SDE~\eqref{eq:reverse-irsde} for multiple steps. 
While this increases the computational cost, it also enables a more accurate restoration of the ground truth. Recently, Refusion~\cite{luo2023refusion} extends the IR-SDE with a U-Net based latent framework to accelerate inference.

Most similar to the IR-SDE is the work of \citet{welker2022speech} and \citet{richter2022speech}, in which a mean-reverting SDE is applied to the speech processing tasks of speech enhancement and speech dereverberation. They use a mean-reverting SDE similar to \eqref{equ:ou} but with a different $\sigma_t$ and a constant $\theta$, i.e. a standard OU process. Also, they did not set the stationary variance condition. In a concurrent work by \citet{welker2022driftrec}, they extend the idea to JPEG artifact removal, where they introduce another version of their SDE with a linear $\theta$ scheduler. As shown in Section~\ref{section:discussion:theta}, both of these are outperformed by our cosine $\theta$ scheduler. Moreover, \citet{welker2022speech, richter2022speech, welker2022driftrec} all use the standard score matching objective, while we introduce an alternative maximum likelihood-based loss function that stabilizes training and improves the restoration performance. Finally, we demonstrate the general applicability of our approach by applying it to six diverse image restoration tasks.

\section{Conclusion}

We have presented a mean-reverting SDE-based method that is applicable to a wide class of image restoration tasks. Importantly, our SDE has a closed-form solution that enables us to compute the ground truth time-dependent score function and to train a neural network to estimate it. In addition, we have proposed a maximum likelihood-based loss objective, which significantly stabilizes the neural network training and consistently improves the restoration performance. The experiments performed on six diverse image restoration tasks demonstrate the wide applicability and highly competitive restoration performance of our proposed approach. Future directions include exploring techniques for optimizing the $\theta$ schedule and sampling procedures which can decrease the computational cost at test time.


\section*{Acknowledgements}
This research was supported by the \emph{Wallenberg AI, Autonomous Systems and Software Program (WASP)} funded by the Knut and Alice Wallenberg Foundation; 
by the project \emph{Deep Probabilistic Regression -- New Models and Learning Algorithms} (contract number: 2021-04301) funded by the Swedish Research Council; and by the 
\emph{Kjell \& M{\"a}rta Beijer Foundation}.
The computations were enabled by the \textit{Berzelius} resource provided by the Knut and Alice Wallenberg Foundation at the National Supercomputer Centre. We also thank Daniel Gedon for providing helpful feedback.


\bibliography{main_paper}
\bibliographystyle{icml2023}



\newpage
\appendix
\onecolumn

\section{Proofs}

\textbf{Proposition 3.1.}
\label{prf:SDE_solution}
    \textit{Suppose that the SDE coefficients in \eqref{equ:ou} satisfy $\sigma_t^2 \, / \, \theta_t = 2 \, \lambda^2$ for all times $t$. Then, given any starting state ${x}(s)$ at time $s < t$, the solution to the SDE is
    \begin{equation}
        {x}(t) = \mu + \bigl({x}(s) - \mu \bigr) \, \expp^{-\bar{\theta}_{s:t}} + \int^t_s \sigma_z \, \expp^{-\bar{\theta}_{z:t}} \diff w(z),
    \end{equation}
    where $\bar{\theta}_{s:t} \coloneqq \int^t_s \theta_z \diff z$ is known and the transition kernel $p({x}(t) \cond {x}(s)) = \mathcal{N}\bigl({x}(t) \cond m_{s:t}({x}(s)), v_{s:t}\bigr)$ is a Gaussian with mean $m_{s:t}$ and variance $v_{s:t}$ given by:
    \begin{equation}
        \begin{split}
        m_{s:t}(x_s) &\coloneqq \mu + ({x}(s) - \mu) \, \expp^{-\bar{\theta}_{s:t}}, \\
        v_{s:t} &\coloneqq \int^t_s \sigma_z^2 \, \expp^{-2\bar{\theta}_{z:t}} \diff z \\
        & \ = \lambda^2 \, \Bigl(1 - \expp^{-2 \, \bar{\theta}_{s:t}}\Bigr).
        \end{split}
    \end{equation}
}


\begin{proof}
Recall the SDE that we want to solve:
\begin{equation}
   \diff {x} = \theta_t \, (\mu - {x}) \diff t + \sigma_t \diff w,
\label{app-equ:ou}
\end{equation}
where $\theta_t$ and $\sigma_t$ are two time-dependent positive functions. Also recall that when the starting state is ${x}(0)$, we substitute $\bar{\theta}_{0:t}$ with $\bar{\theta}_{t}$ for notation simplicity. To solve the SDE above, let us define a surrogate differentiable function
\begin{equation}
    \psi({x}, t) = {x}\, \expp^{{\bar{{\theta}}_t}},
    \label{app-eq:function_phi}
\end{equation}
and by It\^{o}'s formula (in the differential form), we have
\begin{align}
\begin{split}
    \diff \psi({x}, t) 
    &= \frac{\partial \psi}{\partial t}({x}, t)\diff t + \frac{\partial \psi}{\partial {x}}({x}, t) {\mathbf f}({x}, t)\diff t   \\
    &\quad+ \frac{1}{2}\frac{\partial^2 \psi}{\partial {x}^2}({x}, t) g(t)^2 \diff t \\
    &\quad+ \frac{\partial \psi}{\partial {x}}({x}, t) g(t) \, \mathrm{dw}_t.
\end{split}
\label{app-eq:Ito_formula}
\end{align}
By substituting ${\mathbf f}({x}, t)$ and $g(t)$ with the drift and the diffusion functions in (\ref{app-equ:ou}), we obtain
\begin{align}
    \diff \psi({x}, t) = \mu {\theta}_t \expp^{\bar{{\theta}}_t}\diff t + \bm{\sigma}_t \expp^{\bar{{\theta}}_t} \, \mathrm{dw}_t 
    \label{eq:solution_phi}
\end{align}
Note here $\diff{\bar{{\theta}}_t} = \diff \int^t_0 \theta_z \diff z = {\theta}_t$. 
Then we can solve ${x}(t)$ conditioned on ${x}(s)$, as
\begin{align}
\begin{split}
    \psi({x}, t) &- \psi({x}, s) = \int_{s}^t \mu {\theta}_z \expp^{\bar{{\theta}}_z}\diff z + \underbrace{\int_{s}^t\bm{\sigma}_z \expp^{\bar{{\theta}}_z}{\mathrm{dw}(z)}}_{\sim \mathcal{N}\left(0, \ \int_{s}^t\bm{\sigma}_z^2 \expp^{2\bar{{\theta}}_z}{\mathrm{d}z}\right)}.
    \label{eq:integral_xt_x0}
\end{split}
\end{align}
Since that ${{\theta}_t}$ and $\bm{\sigma}_t$ are scalar-valued, we can analytically compute the two integrals above and then obtain
\begin{align}
\begin{split}
    {x}(t) \expp^{{\bar{{\theta}}_t}} - {x}(s) \expp^{{\bar{{\theta}}_s}} = \mu (\expp^{\bar{{\theta}}_t} - \expp^{\bar{{\theta}}_s}) + \int_{s}^t\bm{\sigma}_z \expp^{\bar{{\theta}}_z}{\mathrm{dw}(z)},
    \label{eq:solve_integral_xt_x0}
\end{split}
\end{align}
By dividing $\expp^{{\bar{{\theta}}_t}}$ to both sides, we have
\begin{equation}
    {x}(t) = \mu + \bigl({x}(s) - \mu \bigr) \, \expp^{-\bar{\theta}_{s:t}} + \int^t_s \sigma_z \, \expp^{-\bar{\theta}_{z:t}} \diff w_z,
    \label{app-eqn:sde_solution}
\end{equation}
which is the solution to the given SDE.

Moreover, we show that the integral term $\int^t_s \sigma_z \, \expp^{-\bar{\theta}_{z:t}} \diff w(z)$ in \eqref{app-eqn:sde_solution} is actually a Gaussian noise $\mathcal{N}\left(0, \ \int_{s}^t\bm{\sigma}_z^2 \expp^{-2\bar{{\theta}}_{z:t}}{\mathrm{d}z}\right)$ such that its variance can be solve by
\begin{align}
\begin{split}
    \int_{s}^t\bm{\sigma}_z^2 \expp^{-2\bar{{\theta}}_{z:t}}{\mathrm{d}z} = \frac{\sigma_t^2}{2\theta_t}\expp^0 - \frac{\sigma_s^2}{2\theta_s} \expp^{-2\bar{{\theta}}_{s:t}} = \lambda^2 \, \Bigl(1 - \expp^{-2 \, \bar{\theta}_{s:t}}\Bigr),
    \label{app_eqm:solve_variance}
\end{split}
\end{align}
under the condition of $\sigma_t^2 \, / \, \theta_t = 2 \, \lambda^2$ for all times $t$. By reparameterizing the solution in \eqref{app-eqn:sde_solution} we arrive at the following transition
\begin{equation}
    p({x}(t) \cond {x}(s)) = \mathcal{N}\bigl({x}(t) \cond m_{s:t}({x}(s)), v_{s:t}\bigr),
\end{equation}
where
\begin{equation}
    \begin{split}
    m_{s:t}(x_s) &\coloneqq \mu + ({x}(s) - \mu) \, \expp^{-\bar{\theta}_{s:t}}, \\
    v_{s:t} &\coloneqq \int^t_s \sigma_z^2 \, \expp^{-2\bar{\theta}_{z:t}} \diff z = \lambda^2 \, \Bigl(1 - \expp^{-2 \, \bar{\theta}_{s:t}}\Bigr).
    \end{split}
\end{equation}
Thus we complete the proof.
\end{proof}

\textbf{Proposition 3.2.}
\label{prf:optimum_trajectory}
    \textit{Given an initial state ${x}_0$, for any state ${x}_i$ at discrete time $i > 0$, the optimum reversing solution for ${x}_i \rightarrow {x}_{i-1}$ in IR-SDE is:
    \begin{equation}
    \begin{split}
        {x}_{i-1}^{*} &= \frac{1 - \expp^{-2 \, \bar{\theta}_{i-1}}}{1 - \expp^{-2 \, \bar{\theta}_i}} \expp^{-\theta_i^{'}} ({x}_i - \mu) + \frac{1 - \expp^{-2 \, \theta_i^{'}}}{1 - \expp^{-2 \, \bar{\theta}_i}} \expp^{-\bar{\theta}_{i-1}} ({x}_0 - \mu) + \mu.
    \end{split}
    \end{equation}
}

\begin{proof}

Recall that the transition distribution $p({x}_i \mid {x}_0)$ and $p({x}_i \mid {x}_{i-1}, {x}_0)$ can be known as stated in Proposition (\ref{prop:forward_sde_solution}). Our objective is to minimize the following negative log likelihood (NLL) to obtain a theoretically optimum path for the reverse SDE:
\begin{equation}
\begin{split}
    {x}_{i-1}^{*} = \arg\min_{{x}_{i-1}} \Bigl[ -\log p \bigl({x}_{i-1} \mid {x}_i, {x}_0 \bigr) \Bigr].
    \label{app-eq:cond_prob_nll}
\end{split}
\end{equation}
By using Bayes' rule, we have
\begin{equation}
\begin{split}
    -\log p \bigl({x}_{i-1} \mid {x}_i, {x}_0 \bigr) &= -\log \frac{p({x}_{i} \mid {x}_{i-1}, {x}_0) p({x}_{i-1} \mid {x}_0)}{p({x}_i \mid {x}_0)} \\[.6em]
    & \propto -\log p\bigl({x}_{i} \mid {x}_{i-1}, {x}_0\bigr) - \log p\bigl({x}_{i-1} \mid {x}_0\bigr)
\end{split}
\label{app-eq:cond_prob_bayes}
\end{equation}
where all transitions are tractable. Then we can directly solve the NLL in \eqref{app-eq:cond_prob_nll} by computing its gradient and setting it to be zero:
\begin{equation}
\begin{split}
     \nabla_{{x}_{t-1}^{*}} \left\{-\log p \bigl({x}_{i-1}^{*} \mid {x}_i, {x}_0 \bigr)\right\} 
     & \propto - \nabla_{{x}_{i-1}^{*}}\log p\bigl({x}_{i} \mid {x}_{i-1}^{*}, {x}_0\bigr) - \nabla_{{x}_{i-1}^{*}} \log p\bigl({x}_{i-1}^{*} \mid {x}_0\bigr) \\[.6em]
     & = - \frac{\expp^{-{\theta}_i^{'}} ({x}_i - \mu - ({x}_{i-1}^{*} - \mu)\expp^{-{\theta}_i^{'}})}{1 - \expp^{-2 \, {\theta}_i^{'}}} + \frac{{x}_{i-1}^{*} - \mu - ({x}_0 - \mu)\expp^{-\bar{\theta}_{i-1}}}{1 - \expp^{-2 \, \bar{\theta}_{i-1}}} \\[.6em]
     & = \frac{({x}_{i-1}^{*} - \mu)\expp^{-2{\theta}_i^{'}}}{1 - \expp^{-2 \, {\theta}_i^{'}}} + \frac{({x}_{i-1}^{*} - \mu)}{1 - \expp^{-2 \, \bar{\theta}_{i-1}}} - \frac{({x}_{i} - \mu)\expp^{-{\theta}_i^{'}}}{1 - \expp^{-2 \, {\theta}_i^{'}}} - \frac{({x}_{0} - \mu)\expp^{-\bar{\theta}_{i-1}}}{1 - \expp^{-2 \, \bar{\theta}_{i-1}}} \\[.6em]
     & = \frac{({x}_{i-1}^{*} - \mu)(1 - \expp^{-2 \, \bar{\theta}_{i}})}{(1 - \expp^{-2 \, {\theta}_i^{'}})(1 - \expp^{-2 \, \bar{\theta}_{i-1}})} - \frac{({x}_{i} - \mu)\expp^{-{\theta}_i^{'}}}{1 - \expp^{-2 \, {\theta}_i^{'}}} - \frac{({x}_{0} - \mu)\expp^{-\bar{\theta}_{i-1}}}{1 - \expp^{-2 \, \bar{\theta}_{i-1}}} = 0,
    \label{app-eq:solve_nll}
\end{split}
\end{equation}
where $\theta_i^{'} = \int_{i-1}^i \theta_t dt$. Since \eqref{app-eq:solve_nll} is linear we get
\begin{equation}
    \begin{split}
        {x}_{i-1}^{*} = \frac{1 - \expp^{-2 \, \bar{\theta}_{i-1}}}{1 - \expp^{-2 \, \bar{\theta}_i}} \expp^{-\theta_i^{'}} ({x}_i - \mu) + \frac{1 - \expp^{-2 \, \theta_i^{'}}}{1 - \expp^{-2 \, \bar{\theta}_i}} \expp^{-\bar{\theta}_{i-1}} ({x}_0 - \mu) + \mu,
    \end{split}
    \end{equation}
which completes the proof (the second-order derivative is a positive constant, i.e. ${x}_{i-1}^{*}$ is indeed the optimal point).
\end{proof}

\section{Denoising SDE/ODE for Gaussian Denoising}
\label{app-sec:denoising_sde}

Here we provide details for the Denoising SDE/ODE as it is a special case of the IR-SDE, in the way of denoting $\mu$ with the HR clean image
\begin{equation}
    \mu \coloneqq {x}_0.
    \label{app-eq:denoise_mu}
\end{equation}
Then we have a simplified transition kernel $p({x}_i \cond {x}_0)$ given by
\begin{equation}
    p_{noise}({x}_i|{x}_0) = \mathcal{N}(m_i({x}_0), v_i),
    \label{app-eq:denoise_transition}
\end{equation}
where
\begin{equation}
    m_i := {x}_0, \quad v_i = \lambda^2 \, \Bigl(1 - \expp^{-2 \, \bar{\theta}_{i}}\Bigr).
\end{equation}
Correspondingly, and the optimum path from ${x}_i \rightarrow {x}_{i-1}$ becomes
\begin{equation}
    \begin{split}
        {x}_{i-1}^{*} &= \frac{1 - \mathrm{e}^{-2\bar{{\theta}}_{i-1}}}{1 - \mathrm{e}^{-2\bar{{\theta}}_{i}}} \mathrm{e}^{-{{\theta}}_{i}} ({x}_i - {x}_0) + {x}_0.
    \end{split}
    \label{app-eq:optimal_denoising_trajectory}
\end{equation}
Recall the reverse-time version of the IR-SDE:
\begin{equation}
    \diff {x} = \big[ \theta_t \, (\mu - {x}) - \sigma_t^2 \, \nabla_{{x}} \log p_t({x}) \big] \diff t + \sigma_t \diff \hat{w},
    \label{app-eq:reverse-irsde}
\end{equation}
and the sampling strategy of ${x}_i$:
\begin{equation}
    {x}_i = m_{i} + \sqrt{v_{i}} \, \epsilon_t, \quad  \epsilon_t \sim \mathcal{N}(0, I).
\label{app_eq:sampling_xt}
\end{equation}
We can then approximate $\mu - {x}$ from \eqref{app_eq:sampling_xt} and combine it with \eqref{eq:noise2score} to rewrite (\ref{app-eq:reverse-irsde}) to the Denoising SDE:
\begin{equation}
    {\mathrm d}{x} = -\frac{1}{2}\bm{\sigma}(t)^2 (1 + \mathrm{e}^{-2\bar{{\theta}}_t}) \nabla_{{x}_t}\log p_t({x}_t){\mathrm d}t + \bm{\sigma}(t){\mathrm d}\bar{\mathbf w}.
    \label{app-eq:denoising-sde-simple}
\end{equation}

In addition, \cite{song2021score} states that there exists a deterministic process that shares the same marginal probability densities as the IR-SDE. Once we have the score, we can also recover images through a deterministic trajectory, as the \textit{probability flow ODE}~\cite{song2021score}:
\begin{equation}
    \diff {x} = \Bigl[\theta_t \, (\mu - {x}) - \frac{1}{2} \, \sigma_t^2 \, \nabla_{{x}}\log p_t({x}) \Bigr]\diff t.
    \label{app-eq:reverse-irode}
\end{equation}
In this denoising case, the corresponding ODE for \eqref{app-eq:denoising-sde-simple} is
\begin{equation}
    {\mathrm d}{x} = -\frac{1}{2}\bm{\sigma}_t^2 \mathrm{e}^{-2\bar{{\theta}}_t} \nabla_{{x}}\log p_t({x}){\mathrm d}t.
    \label{app-eq:denoising-ode-simple}
\end{equation}
Moreover, once we know the real noise level $\sigma_{\mathrm{real}}$ of an image, we can easily derive a appropriate timestep $t^*$ such that the variance of $p_{noise}({x}_i|{x}_0)$ happens to be the noise level:
\begin{equation}
    \sigma_{\mathrm{real}}^2 = v_t = \lambda^2 \, \Bigl(1 - \expp^{-2 \, \bar{\theta}_{t}}\Bigr).
\end{equation}
By solving the $\bar{{\theta}}_t$ we have
\begin{equation}
    t^* = \arg\min_{t} \| \bar{{\theta}}_t - \frac{1}{2\Delta t} \log (1 - \frac{\sigma_{\mathrm{real}}^2}{\lambda^2}) \|. 
\end{equation}
where $\Delta t$ denotes the time interval. Based on it, our method is able to process arbitrary noise levels and can start denoising from middle states, which is more practical and improves sample efficiency.

\section{Relationship between Maximum Likelihood Objective and DDPM}
\label{app-sec:mlo_ddpm}

To further illustrate the maximum likelihood objective, here we also apply it to the Denoising Diffusion Probabilistic Models (DDPM)~\cite{ho2020denoising} to mathematically show the connection with diffusion models.

Consider the diffusion process in DDPM:
\begin{equation}
    \begin{split}
        q(x_t \mid x_{t-1}) &= \mathcal{N}(x_t; \sqrt{1 - \beta_t} x_{t-1}, \beta_t I), \\
        q(x_t \mid x_0) &= \mathcal{N}(x_t; \sqrt{\bar{\alpha}_t} x_0, (1 - \bar{\alpha}_t) I),
    \end{split}
\end{equation}
where $\alpha_t = 1 - \beta_t$ and $\bar{\alpha}_t = \prod_{s=1}^t \alpha_s$. Its reverse transition distribution can be derived from Bayes' rule:
\begin{equation}
    \begin{split}
        q(x_{i-1} \mid x_i, x_0) \propto q(x_{i} \mid x_{i-1}, {x}_0) q(x_{{i-1}} \mid { x}_0). \\
    \end{split}
\end{equation}
Then we can minimize its negative log-likelihood (NLL) to get the optimal $x_{t-1}^*$, as our Proposition 3.2. More specifically, set the gradient of the NLL to zero:
\begin{equation}
    \begin{split}
        & \nabla_{x_{t-1}^{*}} \left\{-\log q \bigl(x_{i-1}^{*} \mid x_i, x_0 \bigr)\right\} \\
        & = \frac{1 - \bar{\alpha}_t}{\beta_t (1 - \bar{\alpha}_{t-1})} x_{t-1}^{*} - (\frac{\sqrt{\alpha_t}}{\beta_t} x_t + \frac{\sqrt{\bar{\alpha}_{t-1}}}{1 - \bar{\alpha}_{t-1}} x_0) = 0.
    \end{split}
\end{equation}
Thus, $x_{t-1}^{*}$ has an optimal value that minimizes the NLL:
\begin{equation}
    x_{t-1}^{*} = \frac{\sqrt{\alpha_t}(1 - \bar{\alpha}_{t-1})}{1 - \bar{\alpha}_t} x_t + \frac{\sqrt{\bar{\alpha}_{t-1}} \beta_t}{1 - \bar{\alpha}_t} x_0,
\end{equation}
which is exactly the reverse mean of DDPM (Eq. (7) in their paper), that guarantees the learning for reverse process.

\section{Additional Implementation Details}
\label{app-sec:implementation}
For all experiments, we use the same noise network: a U-Net similar to DDPM~\cite{chung2022diffusion} but removes all group normalization layers and self-attention layers for inference efficiency. To handle different image sizes, we pad all inputs to make sure outputs could have the same sizes as inputs. The CNN-baseline uses the same network but directly input the low-quality image and output the high-quality image.
The stationary variance $\lambda^2$ is set to 10 (over 255) and we use only 100 steps for all experiments since the forward process of IR-SDE could be non-Markov (conditioning on $x_0$), as in DDIM~\cite{song2021denoising}. 

For most tasks, we set the training patch-size to be $128 \times 128$ and use a batch size of 16. We use Adam~\cite{kingma2014adam} optimizer with parameters $\beta_1 = 0.9$ and $\beta_2=0.99$. The total training steps are fixed to 500 thousand and the initial learning rate set to $10^{-4}$ and decays half per 200 thousand iterations. All of our models are trained on an A100 GPU with 40GB memory for about 1.5 days ($400\thinspace000$ iterations), the same as for the CNN-baseline.

In addition, we define our $\theta$ schedule to be the flipped version to the cosine noise schedule in~\cite{nichol2021improved}:
\begin{equation}
    \theta_t = 1 - \frac{f(t)}{f(0)}, \quad f(t) = \cos \left( \frac{t/T + s}{1 + s} \cdot \frac{\pi}{2} \right)^2,
\end{equation}
where $s=0.008$ as the same as in ~\cite{nichol2021improved}. This cosine $\theta$ schedule is also visually shown in Figure~\ref{fig:cosine_variance}. Once the $\theta$ function is determined, we can compute the corresponding diffusion coefficient $\sigma_t$ by the following stationary condition $\frac{\sigma_t^2}{2 \, \theta_t} \,  = \, \lambda^2$.
Practically, we approximate $\bar{\theta}_t$ using a discrete form as $\bar{\theta}_t \approx \sum_{i=1}^t \theta_i \Delta t$. To alleviate the over-smooth problem as mentioned in Section~\ref{sec:limitations}, we let the exponential term $\expp^{- \, \bar{\theta}_T}$ to be a smaller value $\delta=0.005$ instead of zero and then $\Delta t$ can also be computed by $\Delta t = -\log \delta / \sum_{i=1}^T \theta_i$.

\section{Additional Experimental Results}
\label{app-sec:results}

Here we show more detailed results for each task. Note that all reported results of the comparison methods are obtained from using their official codes and pretrained models.

\textbf{Comparison of losses.}
We first give the final quantitative results of learning deraining task with the proposed maximum likelihood loss~\eqref{eq:nll_objective} and with the noise matching loss~\eqref{eq:noise_objective} in Table~\ref{table:cmp_loss1}. The results show that the maximum likelihood significantly improves the performance over all criteria, which is consistent to Section \ref{sec:discuss_objective} and the result in Figure~\ref{fig:loss_curves}. 

\textbf{Additional quantitative results.}
Here we give the quantitative results of dehazing in Table~\ref{app-table:dehaze}. Here we only compare with the CNN-baseline since DDRM~\cite{kawar2022denoising} requires the degradation parameters to be known, which limits its application on image dehazing.
The comprehensive results of image denoising on three different test sets over different noise levels are given by Tables~\ref{app-table:denoising_mcmaster}, \ref{app-table:denoising_kodak24}, and~\ref{app-table:denoising_cbsd68}. Note we add a SOTA denoising method KBNet~\cite{zhang2023kbnet} on the CBSD68 dataset. The proposed Denoising-ODE has the best perceptual performance for all scenes. 

\textbf{Model complexities.}
We also provide the comparison of computational efficiency and model complexity in Table~\ref{app-table:complexity}. our model only slightly increases the parameters and flops of CNN-baseline, while other SOTA methods have to rely on huge computation operations (MAXIM) or complex network structures (KBNet) to improve their performances. We also need to mention that the reverse process involves repeated network evaluations that increases the inference time and computational cost, which is a common limitation of diffusion models. But for training, we only need to sample noises and learn them, which usually converges quickly. 

\begin{figure}[ht]
\begin{center}
\centerline{\includegraphics[width=.9\columnwidth]{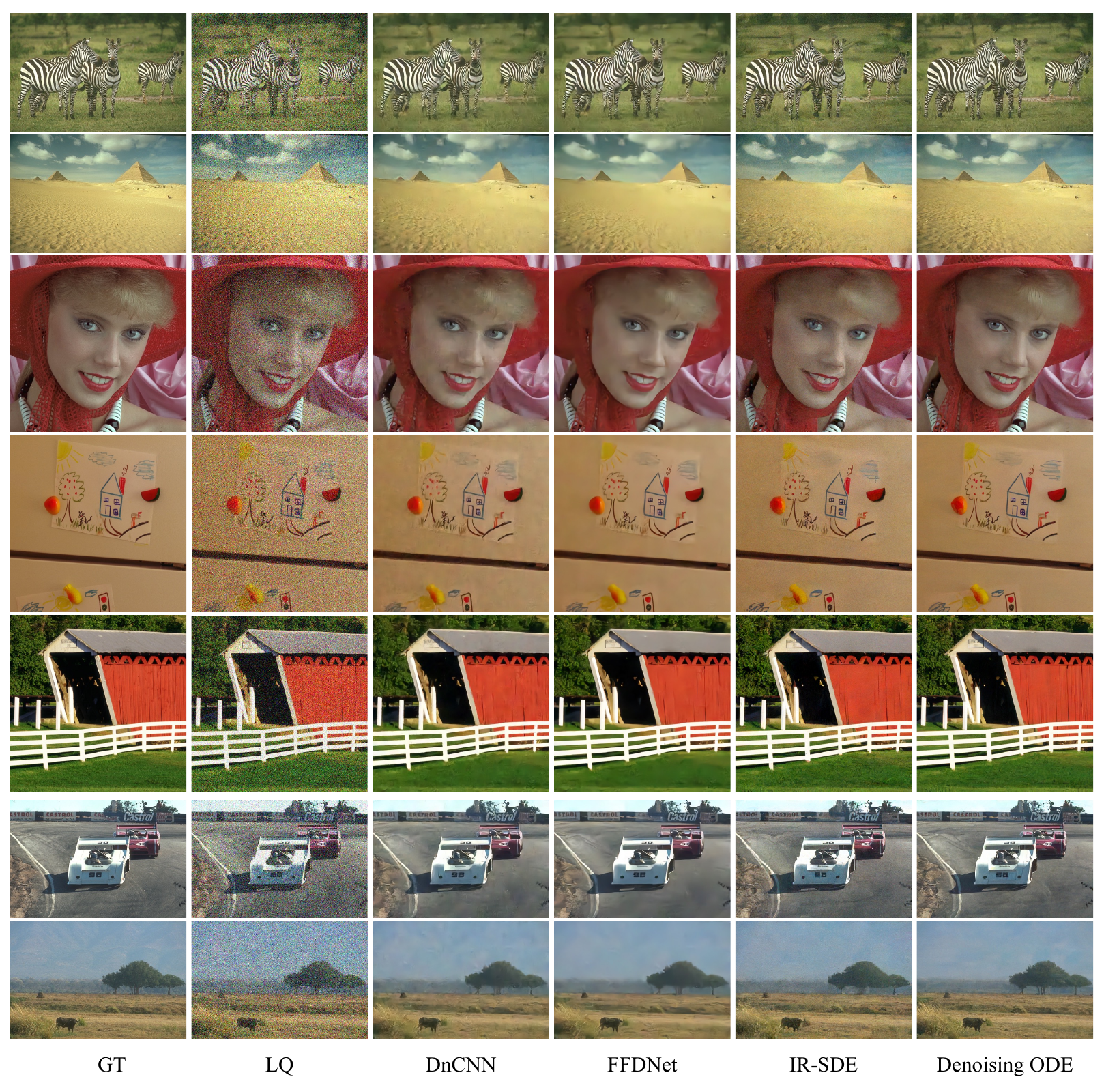}}\vspace{-2.0mm}
\caption{Visual results of our methods with other denosing approaches on Denoising dataset. The noise level $\sigma=50$.}
\label{app-fig:denosing_results}
\end{center}
\vskip -0.2in
\end{figure}

\begin{figure}[ht]
\begin{center}
\centerline{\includegraphics[width=.9\columnwidth]{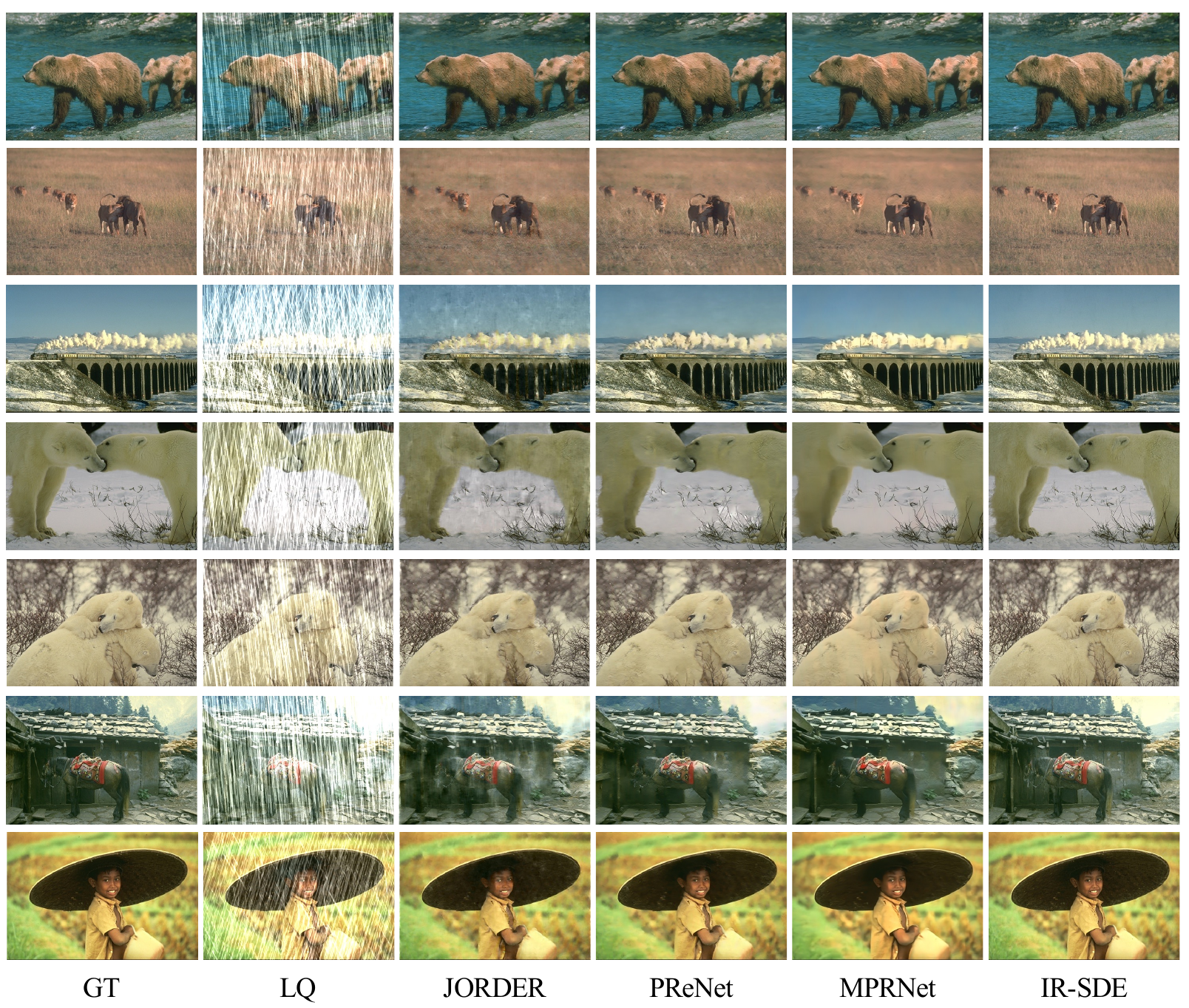}}\vspace{-2.0mm}
\caption{Visual results of our methods with other deraining approaches on Rain100H dataset.}
\label{app-fig:deraining_results}
\end{center}
\vskip -0.2in
\end{figure}

\begin{figure}[ht]
\begin{center}
\centerline{\includegraphics[width=.9\columnwidth]{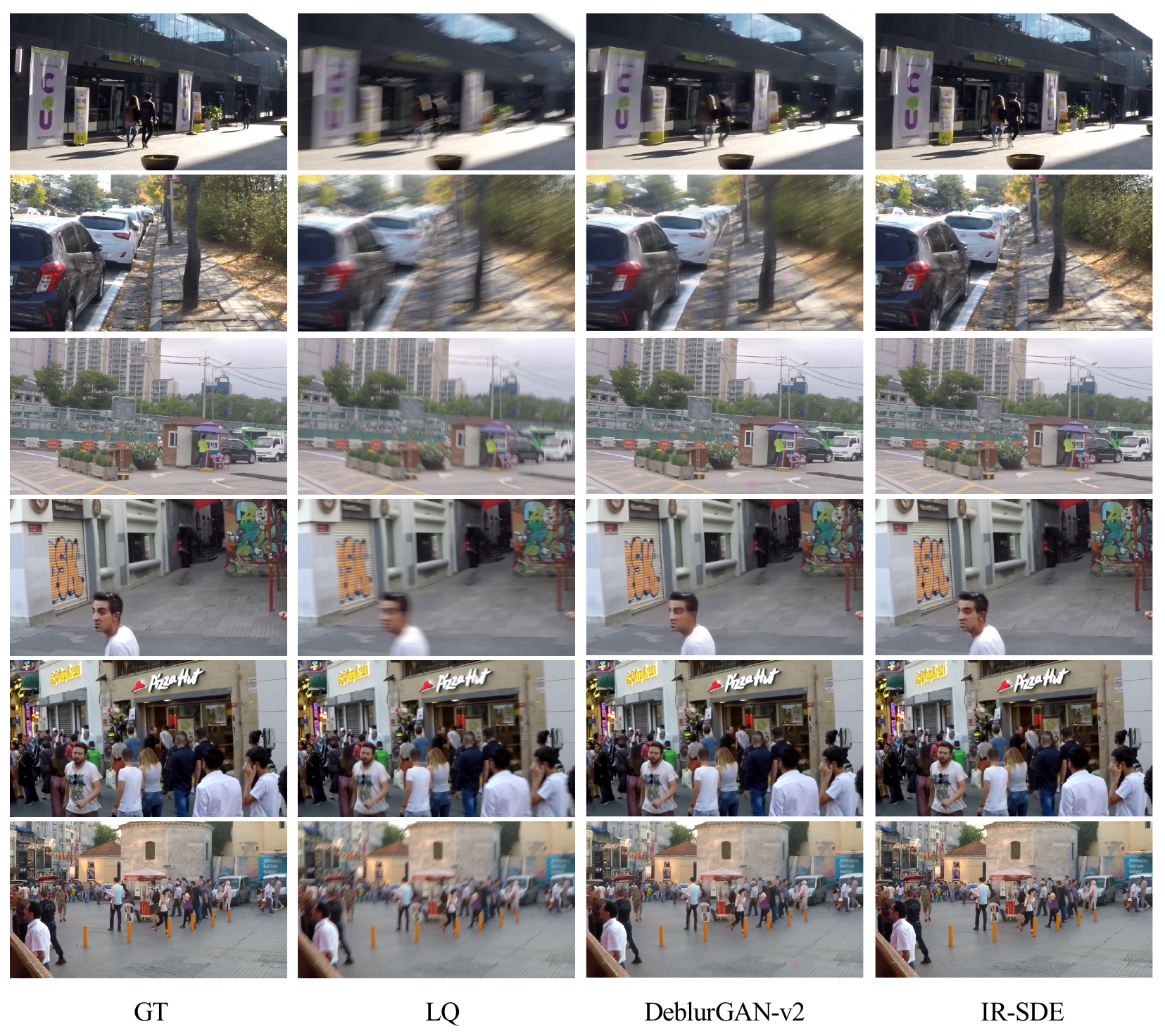}}\vspace{-2.0mm}
\caption{Visual results of on Deblurring task.}
\label{app-fig:deblurring_results}
\end{center}
\vskip -0.2in
\end{figure}

\begin{figure}[ht]
\begin{center}
\centerline{\includegraphics[width=.75\columnwidth]{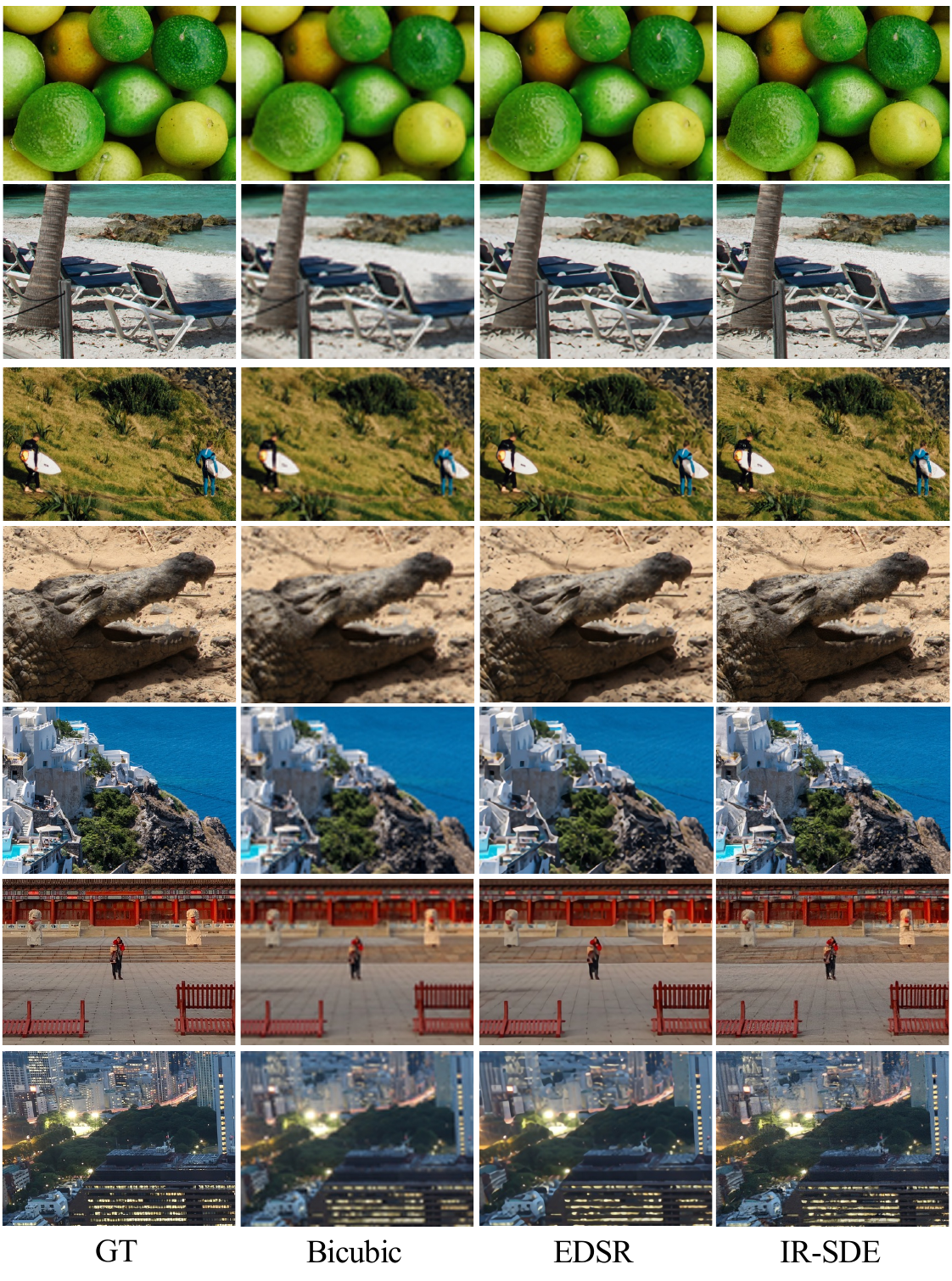}}\vspace{-2.0mm}
\caption{Visual results of our methods on super-resolution task.}
\label{app-fig:sr_results}
\end{center}
\vskip -0.2in
\end{figure}

\begin{figure}[ht]
\begin{center}
\centerline{\includegraphics[width=.75\columnwidth]{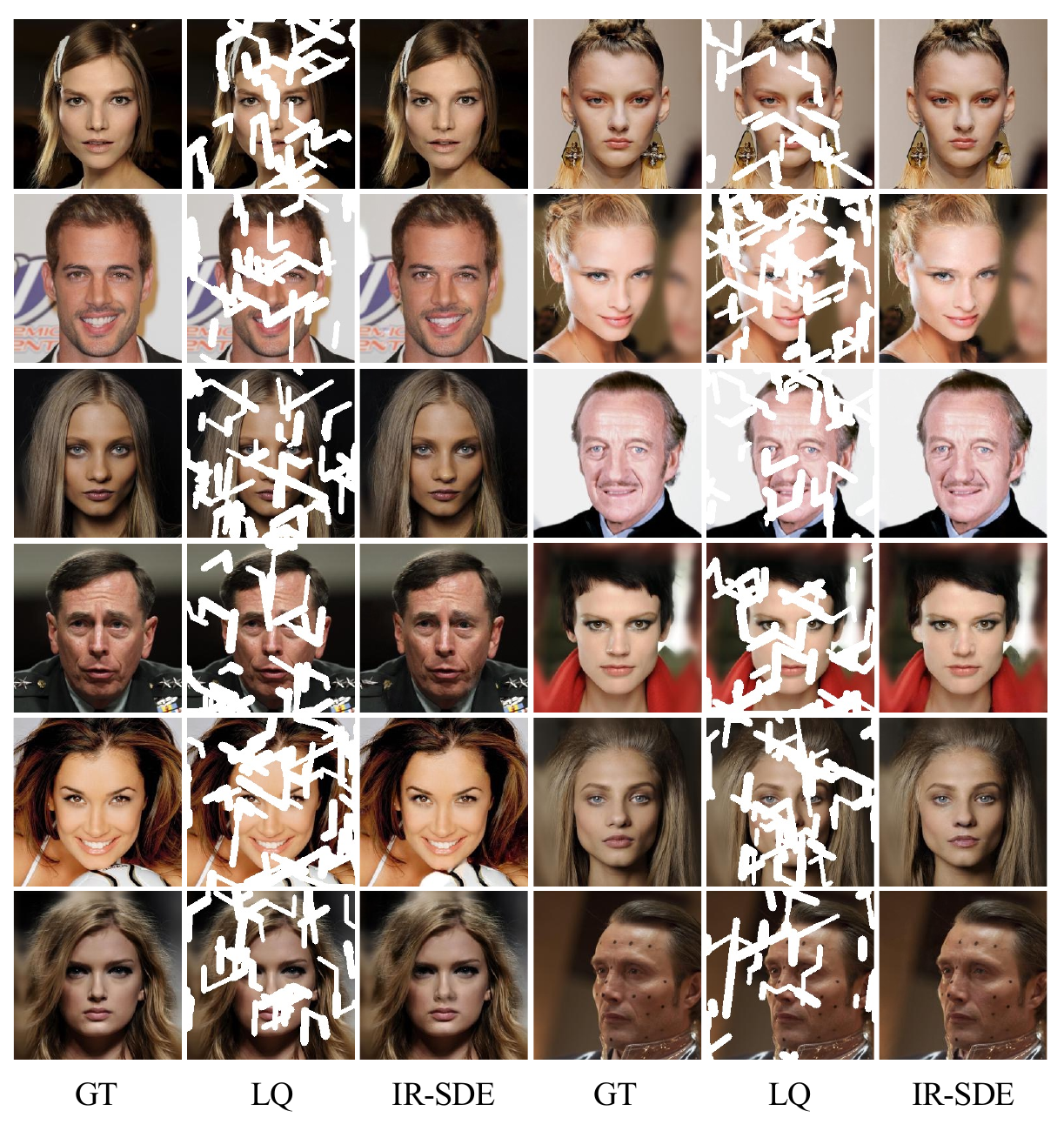}}\vspace{-2.0mm}
\caption{Visual results of our method on inpainting task.}
\label{app-fig:inpainting_results}
\end{center}
\vskip -0.2in
\end{figure}

\begin{figure}[ht]
\begin{center}
\centerline{\includegraphics[width=.8\columnwidth]{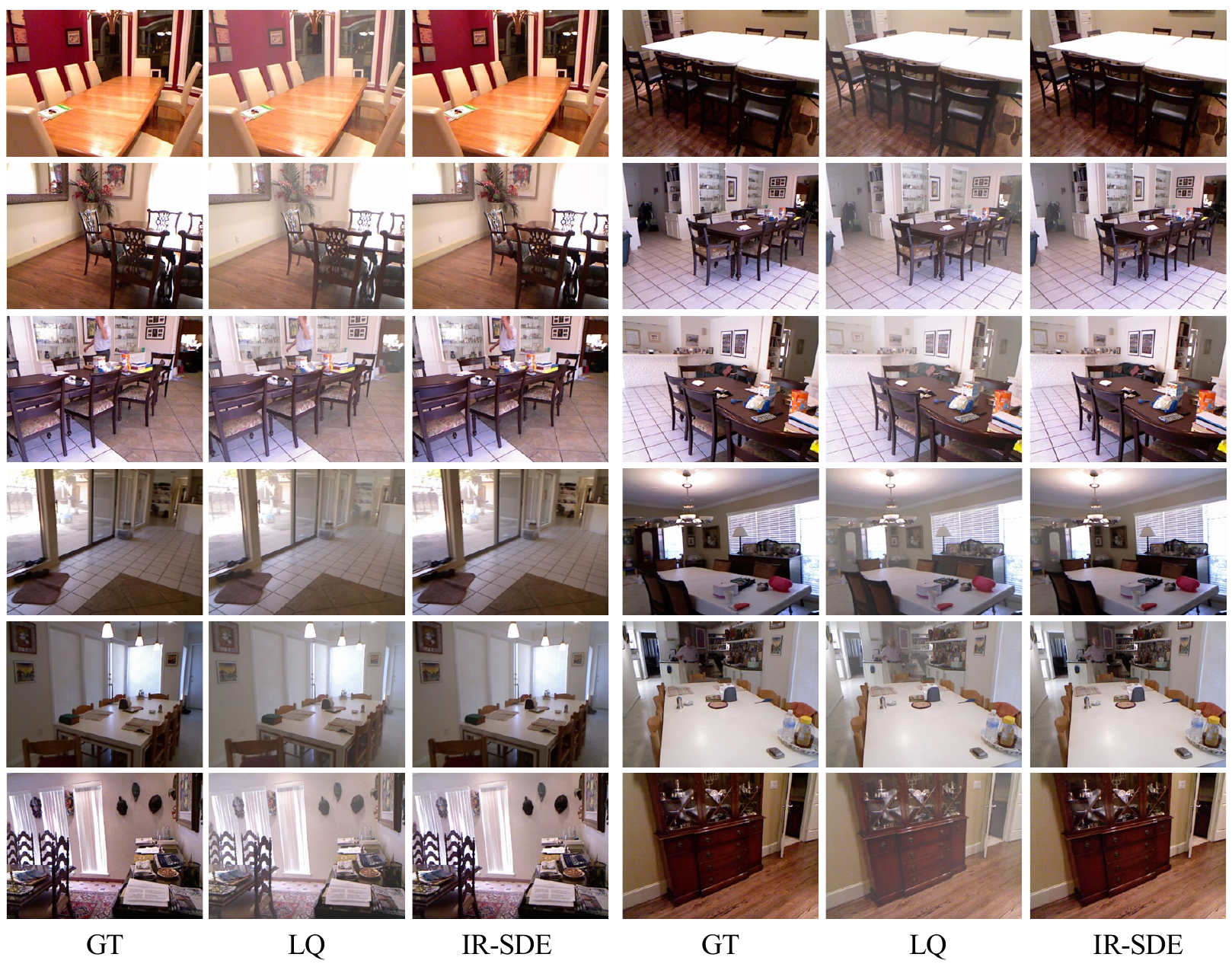}}\vspace{-2.0mm}
\caption{Visual results of our method on dehazing task.}
\label{app-fig:dehazing_results}
\end{center}
\vskip -0.2in
\end{figure}

\textbf{Additional qualitative results.}
We provide additional results on each task. Specifically, Figure~\ref{app-fig:denosing_results},
Figure~\ref{app-fig:deraining_results}, Figure~\ref{app-fig:deblurring_results}, Figure~\ref{app-fig:sr_results}, Figure~\ref{app-fig:inpainting_results}, and Figure~\ref{app-fig:dehazing_results} illustrate visual results on denosing, deraining, deblurring, super-resolution, inpainting, and dehazing, respectively. In most tasks, the results produced by our method are sharper and more realistic. Please zoom in for the best view.

\vskip 1.0in


\begin{table}[ht]

\begin{minipage}[h]{1.\linewidth}
\caption{Analysis of different losses on deraining task.}
\label{table:cmp_loss1}
\begin{center}
\begin{small}
\begin{sc}
\resizebox{.85\linewidth}{!}{
\begin{tabular}{lcccccccc}
\toprule
\multirow{2}{*}{Method} &  \multicolumn{4}{c}{Rain100H dataset} & \multicolumn{4}{c}{Rain100L dataset}  \\ \cmidrule(lr){2-5} \cmidrule(lr){6-9}
& PSNR$\uparrow$ & SSIM$\uparrow$ & LPIPS$\downarrow$ & FID$\downarrow$ & PSNR$\uparrow$ & SSIM$\uparrow$ & LPIPS$\downarrow$ & FID$\downarrow$    \\
\midrule
Maximum Likelihood Loss  & \textbf{30.75} & \textbf{0.9027} & \textbf{0.048} & \textbf{19.76} & \textbf{38.30} & \textbf{0.9805} & \textbf{0.014} & \textbf{7.94} \\
Noise Matching Loss  & 23.59 & 0.7373 & 0.221 & 91.49 & 31.81 & 0.9313 & 0.107 & 52.64 \\
\bottomrule
\end{tabular}
}
\end{sc}
\end{small}
\end{center}
\vskip -0.1in
\end{minipage}

\begin{minipage}[h]{1.\linewidth}
\caption{Quantitative results of dehazing on the SOTS indoor dataset.}
\label{app-table:dehaze}
\vskip 0.05in
\begin{center}
\begin{small}
\begin{sc}
\resizebox{.45\linewidth}{!}{
\begin{tabular}{lcccc}
\toprule
\multirow{2}{*}{Method} &  \multicolumn{2}{c}{Distortion} & \multicolumn{2}{c}{Perceptual}  \\ \cmidrule(lr){2-3} \cmidrule(lr){4-5}
&  PSNR$\uparrow$ & SSIM$\uparrow$ & LPIPS$\downarrow$ & FID$\downarrow$ \\
\midrule
CNN-baseline  & 29.78 & 0.9683 & 0.037 & 34.77  \\
Our Method  & \textbf{34.14} & \textbf{0.9886} & \textbf{0.012} & \textbf{6.06}  \\

\bottomrule
\end{tabular}
}
\end{sc}
\end{small}
\end{center}
\vskip -0.1in
\end{minipage}

\begin{minipage}[h]{1.\linewidth}
\caption{Comparison of the number of parameters and model computational efficiency.}
\label{app-table:complexity}
\vskip 0.05in
\begin{center}
\begin{small}
\begin{sc}
\resizebox{.55\linewidth}{!}{
\begin{tabular}{lcccc}
\toprule
Method &  MAXIM & KBNet & CNN-baseline & Ours  \\
\midrule
\#Parameters  & 14.1M & 118.5M & 33.8M & 34.2M  \\
Flops  & 216G & 68.7G & 98.0G & 98.3G  \\

\bottomrule
\end{tabular}
}
\end{sc}
\end{small}
\end{center}
\vskip -0.1in
\end{minipage}

\end{table}

\begin{table}[ht]

\begin{minipage}[h]{1.\linewidth}
\caption{Quantitative results of image denoising on McMaster~\cite{zhang2011color} test set.}
\label{app-table:denoising_mcmaster}
\begin{center}
\begin{small}
\begin{sc}
\resizebox{1.\linewidth}{!}{
\begin{tabular}{lcccccccccccc}
\toprule
\multirow{2}{*}{Method} &  \multicolumn{4}{c}{$\sigma=15, t^*=15$} & \multicolumn{4}{c}{$\sigma=25,t^*=22$} & \multicolumn{4}{c}{$\sigma=50,t^*=39$} \\ \cmidrule(lr){2-5} \cmidrule(lr){6-9} \cmidrule(lr){10-13}
&  PSNR$\uparrow$ & SSIM$\uparrow$ & LPIPS$\downarrow$ & FID$\downarrow$ &  PSNR$\uparrow$ & SSIM$\uparrow$ & LPIPS$\downarrow$ & FID$\downarrow$ &  PSNR$\uparrow$ & SSIM$\uparrow$ & LPIPS$\downarrow$ & FID$\downarrow$  \\
\midrule
DnCNN  & 33.45 & 0.9035 & 0.068 & 37.14 & 31.52 & 0.8692 & 0.101 & 59.16 & 28.62 & 0.7986 & 0.173 & 107.31 \\
FFDNet  & 34.66 & \textbf{0.9216} & 0.065 & 39.37 & 32.36 & \textbf{0.8861} & 0.103 & 63.84 & \textbf{29.19} & \textbf{0.8149} & 0.183 & 118.38 \\

CNN-baseline  & 33.51 & 0.8978 & 0.089 & 43.90 & 31.79 & 0.8697 & 0.122 & 66.47 & 29.15 & 0.8122 & 0.160 & 93.68 \\

IR-SDE & 31.95 & 0.8600 & 0.038 & 23.97 & 29.48 & 0.8052 & 0.071 & 44.77 & 27.14 & 0.7549 & 0.151 & 97.53 \\

Denoising-ODE  & \textbf{34.80} & 0.9188 & \textbf{0.036} & \textbf{22.03} & \textbf{32.39} & 0.8791 & \textbf{0.055} & \textbf{34.66} & 29.03 & 0.7911 & \textbf{0.091} & \textbf{63.84} \\
Denoising-SDE  & 31.18 & 0.8195 & 0.049 & 29.21 & 28.98 & 0.7512 & 0.088 & 45.84 & 25.85 & 0.6272 & 0.173 & 92.19 \\

\bottomrule
\end{tabular}
}
\end{sc}
\end{small}
\end{center}
\vskip -0.1in
\end{minipage}

\begin{minipage}[h]{1.\linewidth}
\caption{Quantitative results of image denoising on Kodak24~\cite{franzen1999kodak} test set.}
\label{app-table:denoising_kodak24}
\begin{center}
\begin{small}
\begin{sc}
\resizebox{1.\linewidth}{!}{

\begin{tabular}{lcccccccccccc}
\toprule
\multirow{2}{*}{Method} &  \multicolumn{4}{c}{$\sigma=15, t^*=15$} & \multicolumn{4}{c}{$\sigma=25,t^*=22$} & \multicolumn{4}{c}{$\sigma=50,t^*=39$} \\ \cmidrule(lr){2-5} \cmidrule(lr){6-9} \cmidrule(lr){10-13}
&  PSNR$\uparrow$ & SSIM$\uparrow$ & LPIPS$\downarrow$ & FID$\downarrow$ &  PSNR$\uparrow$ & SSIM$\uparrow$ & LPIPS$\downarrow$ & FID$\downarrow$ &  PSNR$\uparrow$ & SSIM$\uparrow$ & LPIPS$\downarrow$ & FID$\downarrow$  \\
\midrule
DnCNN  & 34.48 & 0.9189 & 0.083 & 21.71 & 32.02 & 0.8763 & 0.129 & 41.96 & 28.83 & 0.7908 & 0.229 & 83.27 \\
FFDNet  & 34.63 & \textbf{0.9215} & 0.085 & 21.57 & 32.13 & \textbf{0.8779} & 0.140 & 44.57 & \textbf{28.98} & \textbf{0.7942} & 0.255 & 89.69 \\

CNN-baseline  & 33.92 & 0.9090 & 0.110 & 24.52 & 32.73 & 0.8666 & 0.161 & 45.81 & 28.89 & 0.7904 & 0.223 & 66.01 \\

IR-SDE & 31.85 & 0.8603 & 0.057 & 15.25 & 28.99 & 0.7772 & 0.106 & 35.19 & 26.83 & 0.7190 & 0.208 & 70.96 \\

Denoising-ODE  & \textbf{34.64} & 0.9184 & \textbf{0.050} & \textbf{13.74} & \textbf{32.14} & 0.8739 & \textbf{0.078} & \textbf{21.47} & 28.75 & 0.7746 & \textbf{0.134} & \textbf{45.96} \\

Denoising-SDE  & 30.89 & 0.8099 & 0.074 & 21.09 & 28.55 & 0.7247 & 0.130 & 36.18 & 25.46 & 0.5788 & 0.249 & 75.33 \\

\bottomrule
\end{tabular}
}
\end{sc}
\end{small}
\end{center}
\vskip -0.1in
\end{minipage}

\begin{minipage}[h]{1.\linewidth}
\caption{Quantitative results of image denoising on CBSD68~\cite{martin2001database} test set.}
\label{app-table:denoising_cbsd68}
\begin{center}
\begin{small}
\begin{sc}
\resizebox{1.\linewidth}{!}{
\begin{tabular}{lcccccccccccc}
\toprule
\multirow{2}{*}{Method} &  \multicolumn{4}{c}{$\sigma=15, t^*=15$} & \multicolumn{4}{c}{$\sigma=25,t^*=22$} & \multicolumn{4}{c}{$\sigma=50,t^*=39$} \\ \cmidrule(lr){2-5} \cmidrule(lr){6-9} \cmidrule(lr){10-13}
&  PSNR$\uparrow$ & SSIM$\uparrow$ & LPIPS$\downarrow$ & FID$\downarrow$ &  PSNR$\uparrow$ & SSIM$\uparrow$ & LPIPS$\downarrow$ & FID$\downarrow$ &  PSNR$\uparrow$ & SSIM$\uparrow$ & LPIPS$\downarrow$ & FID$\downarrow$  \\
\midrule
DnCNN  & \textbf{33.90} & 0.9289 & 0.063 & 25.59 & 31.24 & 0.8830 & 0.109 & 43.51 & 27.95 & \textbf{0.7896} & 0.210 & 84.56 \\
FFDNet  & 33.88 & \textbf{0.9290} & 0.065 & 27.24 & 31.22 & 0.8821 & 0.121 & 49.64 & \textbf{27.97} & 0.7887 & 0.244 & 98.76 \\

KBNet & - & - & - & - & \textbf{31.71} & \textbf{0.8923} & 0.098 & 37.86 & - & - & - & - \\

CNN-baseline  & 33.02 & 0.9139 & 0.098 & 31.99 & 30.74 & 0.8661 & 0.162 & 56.64 & 27.84 & 0.7827 & 0.232 & 78.51 \\

IR-SDE & 31.04 & 0.8708 & 0.055 & 21.56 & 28.09 & 0.7866 & 0.101 & 36.49 & 25.54 & 0.6894 & 0.219 & 97.95 \\

Denoising-ODE  & 33.80 & 0.9251 & \textbf{0.042} & \textbf{16.71} & 31.14 & 0.8777 & \textbf{0.074} & \textbf{28.71} & 27.59 & 0.7733 & \textbf{0.138} & \textbf{50.46} \\
Denoising-SDE  & 30.15 & 0.8270 & 0.078 & 25.32 & 27.65 & 0.457 & 0.131 & 39.25 & 24.37 & 0.5875 & 0.243 & 84.87 \\

\bottomrule
\end{tabular}
}
\end{sc}
\end{small}
\end{center}
\vskip -0.1in
\end{minipage}

\end{table}

\end{document}